\documentclass{article}

\usepackage[final, nonatbib]{neurips_2020}

\usepackage[utf8]{inputenc} 
\usepackage[T1]{fontenc}    
\usepackage{url}            
\usepackage{booktabs}       
\usepackage{amsfonts}       
\usepackage{nicefrac}       
\usepackage{microtype}      
\usepackage{epsfig}
\usepackage{amsmath}
\usepackage{mathtools}
\usepackage{caption}
\usepackage{subcaption}
\usepackage{algorithm}
\usepackage{algpseudocode}
\usepackage{authblk}
\usepackage{bbm}
\usepackage{pifont}
\usepackage{color}

\usepackage{amsmath,amsfonts,bm}

\def\eqref#1{equation~\ref{#1}}

\def\1{\bm{1}}

\def\ry{{\textnormal{y}}}
\def\rz{{\textnormal{z}}}

\def\rvx{{\mathbf{x}}}

\def\rvz{{\mathbf{z}}}

\def\va{{\bm{a}}}

\def\vx{{\bm{x}}}
\def\vy{{\bm{y}}}
\def\vz{{\bm{z}}}

\def\mA{{\bm{A}}}

\def\mX{{\bm{X}}}

\DeclareMathAlphabet{\mathsfit}{\encodingdefault}{\sfdefault}{m}{sl}
\SetMathAlphabet{\mathsfit}{bold}{\encodingdefault}{\sfdefault}{bx}{n}

\usepackage{multirow}
\usepackage{textcomp}
\usepackage{comment}

\newcommand{\etal}{\textit{et al.~}}

\usepackage[colorlinks = true,
            linkcolor = black,
            urlcolor  = black,
            citecolor = black,
            anchorcolor = black]{hyperref}

\newcommand{\MYhref}[3][blue]{\href{#2}{\color{#1}{#3}}}

\title{ContraGAN: Contrastive Learning for\\ Conditional Image Generation}

\author{%
\textbf{Minguk Kang} \hspace{1cm} \textbf{Jaesik Park}\\
  Graduate School of Artificial Intelligence\\
  POSTECH\\
  \texttt{\{mgkang, jaesik.park\}@postech.ac.kr} \\
}

\begin{document}
\maketitle
\begin{abstract}
Conditional image generation is the task of generating diverse images using class label information. Although many conditional Generative Adversarial Networks (GAN) have shown realistic results, such methods consider pairwise relations between the embedding of an image and the embedding of the corresponding label (\emph{data-to-class relations}) as the conditioning losses. In this paper, we propose ContraGAN that considers relations between multiple image embeddings in the same batch (\emph{data-to-data relations}) as well as the data-to-class relations by using a conditional contrastive loss. The discriminator of ContraGAN discriminates the authenticity of given samples and minimizes a contrastive objective to learn the relations between training images. Simultaneously, the generator tries to generate realistic images that deceive the authenticity and have a low contrastive loss. The experimental results show that ContraGAN outperforms state-of-the-art-models by 7.3\% and 7.7\% on Tiny ImageNet and ImageNet datasets, respectively. Besides, we experimentally demonstrate that ContraGAN helps to relieve the overfitting of the discriminator. For a fair comparison, we re-implement twelve state-of-the-art GANs using the PyTorch library. The software package is available at~\MYhref[magenta]{https://github.com/POSTECH-CVLab/PyTorch-StudioGAN}{https://github.com/POSTECH-CVLab/PyTorch-StudioGAN}.
\end{abstract}
\section{Introduction}
Generative Adversarial Networks (GAN)~\cite{Goodfellow2014GenerativeAN} have introduced a new paradigm for realistic data generation. Many approaches have shown impressive improvements in un/conditional image generation tasks~\cite{Radford2016UnsupervisedRL, Arjovsky2017WassersteinG, Miyato2018SpectralNF, Zhang2019SelfAttentionGA, Brock2019LargeSG, Zhang2019ConsistencyRF, Zhao2020ImprovedCR, Wu2019LOGANLO}. The studies on non-convexity of objective landscapes~\cite{Kodali2018OnCA, Li2018OnTL, Nagarajan2017GradientDG} and gradient vanishing problems~\cite{Arjovsky2017WassersteinG, Li2018OnTL, Mao2017LeastSG, Arjovsky2017TowardsPM} emphasize the instability of the adversarial dynamics. Therefore, many approaches have tried to stabilize the training procedure by adopting well-behaved objectives~\cite{Arjovsky2017WassersteinG, Mao2017LeastSG, Lim2017GeometricG} and regularization techniques~\cite{Miyato2018SpectralNF, Zhang2019ConsistencyRF, Gulrajani2017ImprovedTO}. In particular, spectral normalization~\cite{Miyato2018SpectralNF} with a projection discriminator~\cite{Miyato2018cGANsWP} made the first success in generating images of ImageNet dataset~\cite{Deng2009ImageNetAL}. SAGAN~\cite{Zhang2019SelfAttentionGA} shows using spectral normalization on both the generator and discriminator can alleviate training instability of GANs. BigGAN~\cite{Brock2019LargeSG}
dramatically advances the quality of generated images by scaling up the number of network parameters and batch size.

On this journey, conditioning class information for the generator and discriminator turns out to be the secret behind realistic image generation~\cite{Miyato2018cGANsWP, Odena2017ConditionalIS, Siarohin2019WhiteningAC}. ACGAN~\cite{Odena2017ConditionalIS} validates this direction by training a softmax classifier along with the discriminator. ProjGAN~\cite{Miyato2018cGANsWP} utilizes a projection discriminator with probabilistic model assumptions. Especially, ProjGAN shows surprising image synthesis results and becomes the basic model adopted by SNGAN~\cite{Miyato2018SpectralNF}, SAGAN~\cite{Zhang2019ConsistencyRF}, BigGAN~\cite{Brock2019LargeSG}, CRGAN~\cite{Zhang2019ConsistencyRF}, and LOGAN~\cite{Wu2019LOGANLO}. However, GANs with the projection discriminator have overfitting issues, which lead to the collapse of adversarial training~\cite{Brock2017NeuralPE, Wu2019LOGANLO, zhao2020differentiable, karras2020training}. The ACGAN is known to be unstable when the number of classes increases~\cite{Miyato2018cGANsWP, Odena2017ConditionalIS}.

In this paper, we propose a new conditional generative adversarial network framework, namely \textit{Contrastive Generative Adversarial Networks} (ContraGAN). Our approach is motivated by an interpretation that ACGAN and ProjGAN utilize \emph{data-to-class} relation as the conditioning losses. Such losses only consider relations between the embedding of an image and the embedding of the corresponding label. In contrast, ContraGAN is based on a conditional contrastive loss (2C loss) to consider \emph{data-to-data} relations in the same batch. ContraGAN pulls the multiple image embeddings closer to each other when the class labels are the same, but it pushes far away otherwise. In this manner, the discriminator can capture not only \emph{data-to-class} but also \emph{data-to-data} relations between samples.

We perform image generation experiments on CIFAR10~\cite{Krizhevsky2009LearningML}, Tiny ImageNet~\cite{Tiny}, and ImageNet~\cite{Deng2009ImageNetAL} datasets using various backbone architectures, such as DCGAN~\cite{Radford2016UnsupervisedRL}, ResGAN~\cite{He_2016_CVPR, Gulrajani2017ImprovedTO}, and BigGAN~\cite{Brock2019LargeSG} equipped with spectral normalization~\cite{Miyato2018SpectralNF}. Through exhaustive experiments, we verify that the proposed ContraGAN improves the state-of-the-art-models by 7.3\% and 7.7\% on Tiny ImageNet and ImageNet datasets respectively, in terms of Frechet Inception Distance~(FID)~\cite{ttur2010inline}. Also, ContraGAN gives comparable results (1.3\% lower FID) on CIFAR10 with the art model~\cite{Brock2019LargeSG}. Since ContraGAN can learn plentiful data-to-data relations from a properly sized batch, it reduces FID significantly \emph{without hard negative and positive mining}. Furthermore, we experimentally show that 2C loss alleviates the overfitting problem of the discriminator. In the ablation study, we demonstrate that ContraGAN can benefit from consistency regularization~\cite{Zhang2019ConsistencyRF} that uses data augmentations.

In summary, the contributions of our work are as follows:
\begin{itemize}
\item{We propose novel Contrastive Generative Adversarial Networks (ContraGAN) for conditional image generation. ContraGAN is based on a novel conditional contrastive loss (2C loss) that can learn both data-to-class and data-to-data relations.}
\item{We experimentally demonstrate that ContraGAN improves state-of-the-art-results by 7.3\% and 7.7\% on Tiny ImageNet and ImageNet datasets, respectively. ContraGAN also helps to relieve the overfitting problem of the discriminator.}
\item ContraGAN shows favorable results without data augmentations for consistency regularization. If consistency regularization is applied, ContraGAN can give superior image generation results.
\item We provide implementations of twelve state-of-the-art GANs for a fair comparison. Our implementation of the prior arts for CIFAR10 dataset achieves even better performances than FID scores reported in the original papers.
\end{itemize}
\section{Background}
\subsection{Generative Adversarial Networks }
Generative adversarial networks (GAN)~\cite{Goodfellow2014GenerativeAN} are implicit generative models that use a generator and a discriminator to synthesize realistic images.
While the discriminator ($D$) should distinguish whether the given images are synthesized or not, the generator ($G$) tries to fool the discriminator by generating realistic images from noise vectors. The objective of the adversarial training is as follows:
\begin{align*}
    \min_{G}\max_{D}\mathbb{E}_{\rvx \sim {p_{\mathrm{real}}(\rvx)}}[\log(D(\rvx))] + \mathbb{E}_{\rvz \sim {p(\rvz)}}[\log(1 - D(G(\rvz)))],
    \label{eq:eq1}\tag{1}
\end{align*}
where $p_{\mathrm{real}}(\rvx)$ is the real data distribution, and $p_{\rvz}(\rvz)$ is a predefined prior distribution, typically multivariate Gaussian. Since the dynamics between the generator and discriminator is unstable, and it is hard to achieve the Nash equilibrium~\cite{Nash1951NONCOOPERATIVEG}, there are many objective functions~\cite{Arjovsky2017WassersteinG, Mao2017LeastSG, Lim2017GeometricG, Nowozin2016fGANTG} and regularization techniques~\cite{Miyato2018SpectralNF, Zhang2019ConsistencyRF, Gulrajani2017ImprovedTO, Brock2017NeuralPE} to help networks to converge to a proper equilibrium.
 
\begin{figure}[t]
    {
     \centering
     \begin{subfigure}[b]{0.29\textwidth}
         \centering
         \includegraphics[width=\textwidth]{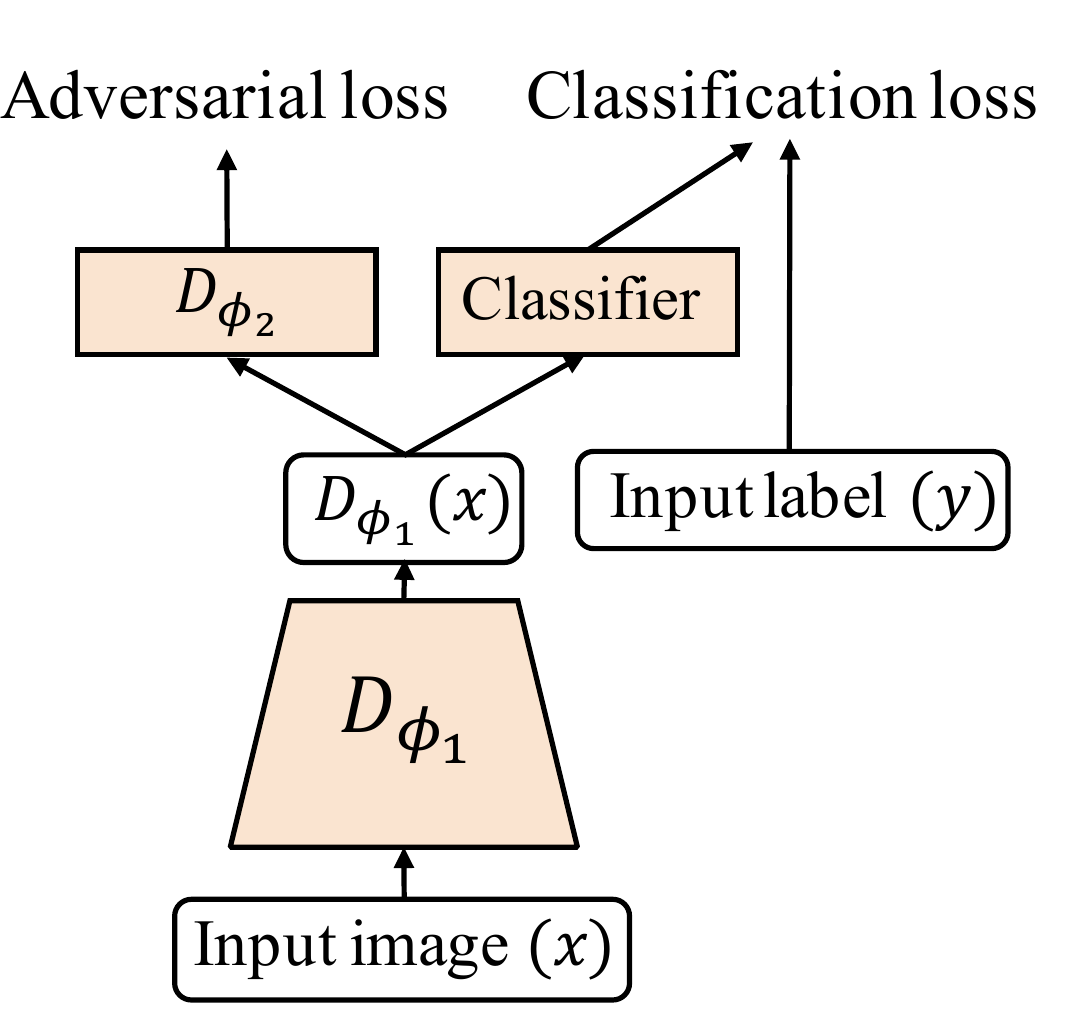}
         \caption{ACGAN~\cite{Odena2017ConditionalIS}}
         \label{fig:fig2a}
     \end{subfigure}
     \begin{subfigure}[b]{0.32\textwidth}
         \centering
         \includegraphics[width=\textwidth]{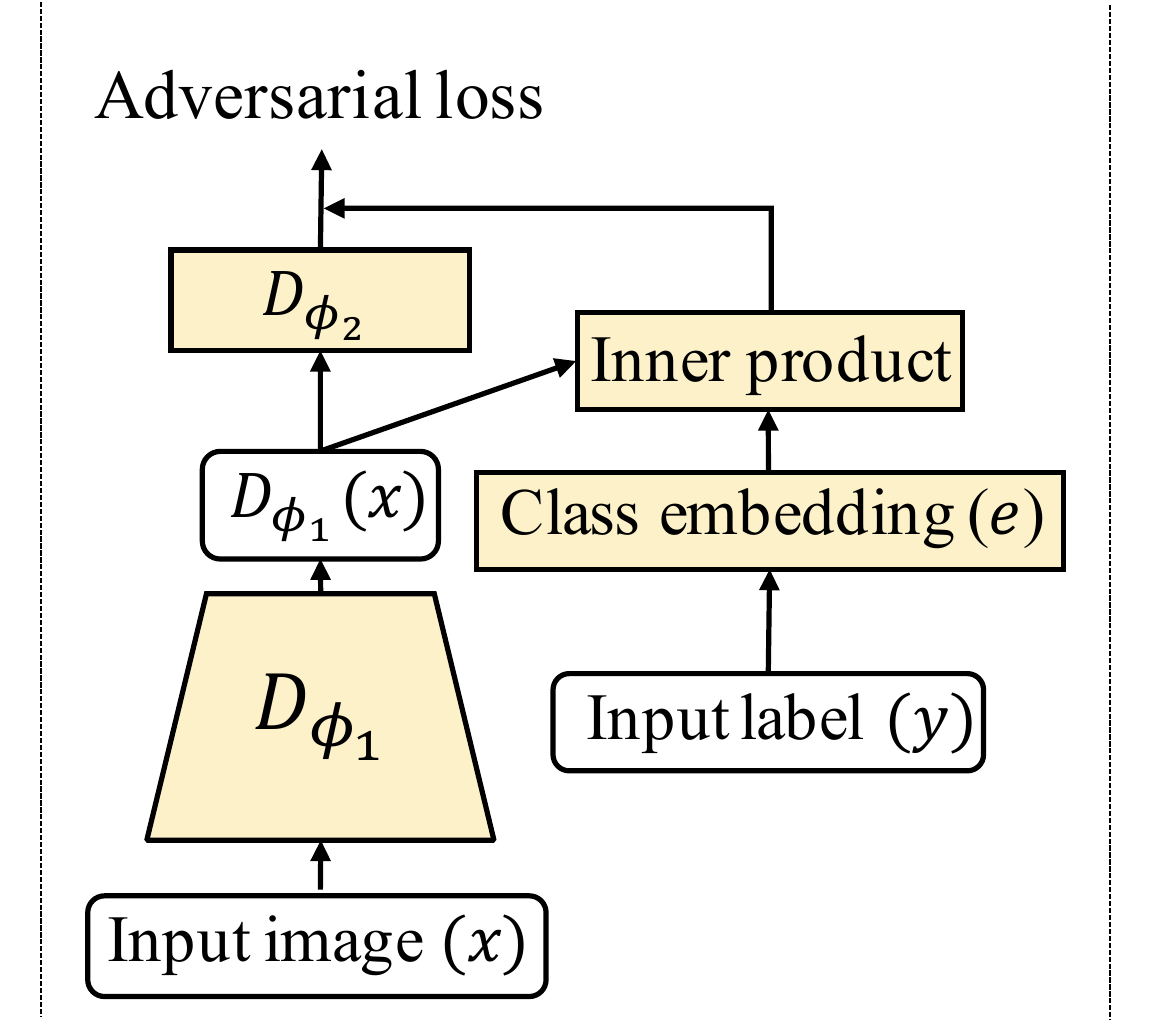}
         \caption{ProjGAN~\cite{Miyato2018cGANsWP}}
         \label{fig:fig2b}
     \end{subfigure}
     \begin{subfigure}[b]{0.37\textwidth}
         \centering
         \includegraphics[width=\textwidth]{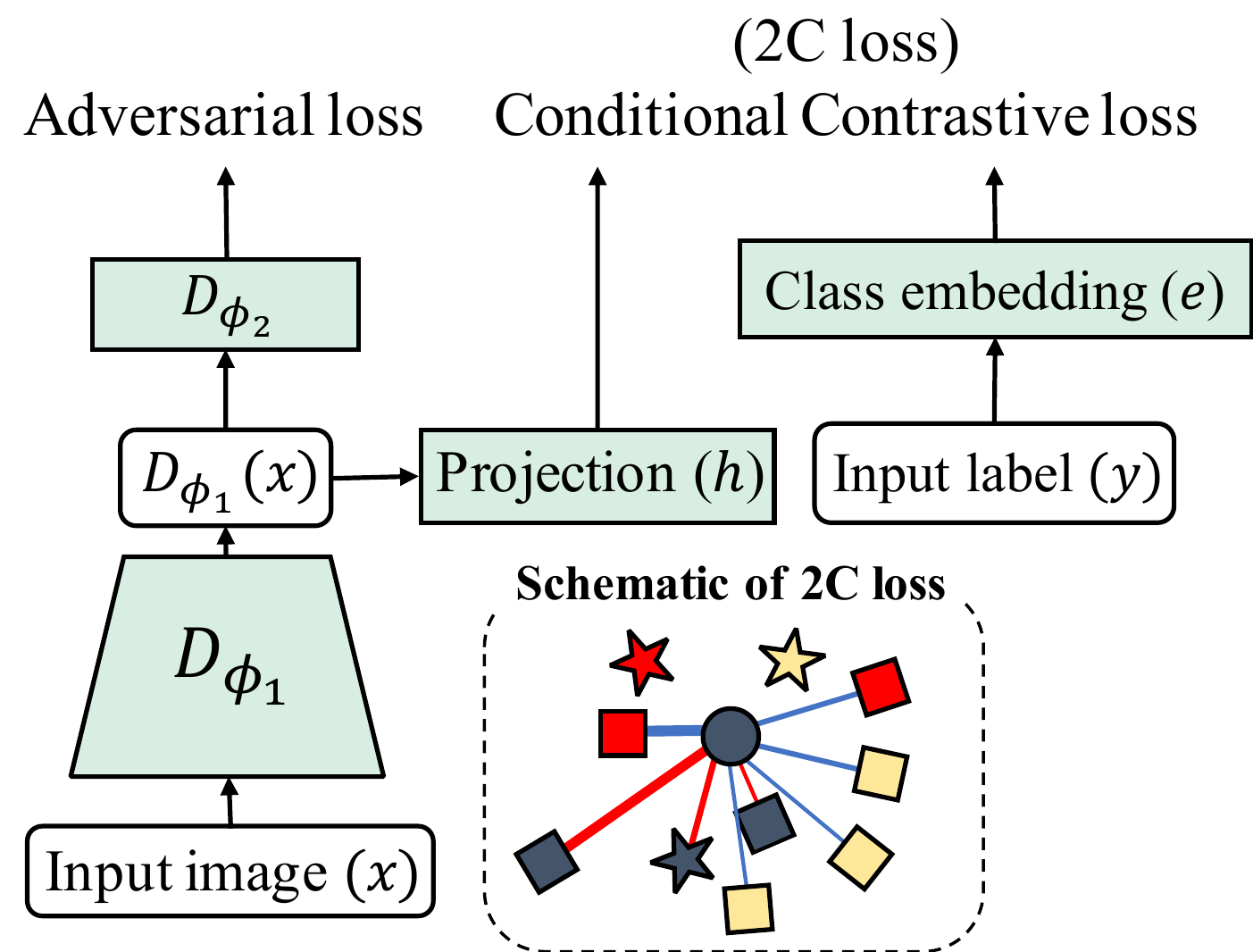}
         \caption{ContraGAN (Ours)}
         \label{fig:fig2c}
     \end{subfigure}
    \caption{Schematics of discriminators of three conditional GANs. (a) ACGAN~\cite{Odena2017ConditionalIS} has an auxiliary classifier to guide the generator to synthesize well-classifiable images. (b) ProjGAN~\cite{Miyato2018cGANsWP} improves ACGAN by adding the inner product of an embedded image and the corresponding class embedding. (c) Our approach extends ACGAN and ProjGAN with a conditional contrastive loss (2C loss). ContraGAN considers multiple positive and negative pairs in the same batch. ContraGAN utilizes 2C loss to update the generator as well.}
    }
\end{figure}

\subsection{Conditional GANs}
\label{sec:conditional_GAN}
One of the widely used strategies to synthesize realistic images is utilizing class label information.
Early approaches in this category are conditional variational auto-encoder (CVAE)~\cite{Sohn2015LearningSO} and conditional generative adversarial networks~\cite{Mirza2014ConditionalGA}. These approaches concatenate a latent vector with the label to manipulate the semantic characteristics of the generated image. Since DCGAN~\cite{Radford2016UnsupervisedRL} demonstrated high-resolution image generation, GANs utilizing class label information has shown advanced performances~\cite{Brock2019LargeSG, Zhang2019ConsistencyRF, Wu2019LOGANLO, Zhao2020ImprovedCR}. 

The most common approach of conditional GANs is to inject label information into the generator and discriminator. ACGAN~\cite{Odena2017ConditionalIS} attaches an auxiliary classifier on the top of convolutional layers in the discriminator to distinguish the classes of images. An illustration of ACGAN is shown in Fig.~\ref{fig:fig2a}. ProjGAN~\cite{Miyato2018cGANsWP} points out that ACGAN is likely to generate easily classifiable images, and the generated images are not diverse. ProjGAN proposes a projection discriminator to relieve the issues (see Fig.~\ref{fig:fig2b}). However, these approaches do not explicitly consider data-to-data relations in the training phase. Besides, the recent study by Wu~\etal~\cite{Wu2019LOGANLO} discovers that BigGAN with the projection discriminator~\cite{Brock2019LargeSG} still suffers from the discriminator's overfitting and training collapse problems.
\section{Method}
We begin with analyzing that conditioning functions of ACGAN and ProjGAN can be interpreted as pair-based losses that look at only data-to-class relations of training examples~(Sec.~\ref{sec:cGAN}). Then, in order to consider both data-to-data and data-to-class relations, we devise a new conditional contrastive loss (2C loss)~(Sec.~\ref{sec:2c_loss}). Finally, we propose Contrastive Generative Adversarial Networks (ContraGAN) for conditional image generation~(Sec.~\ref{sec:contraGAN}).

\subsection{Conditional GANs and Data-to-Class Relations}
\label{sec:cGAN}
\begin{figure}[t]
     \centering
     \begin{subfigure}[b]{0.16\textwidth}
         \centering
         \includegraphics[width=\textwidth]{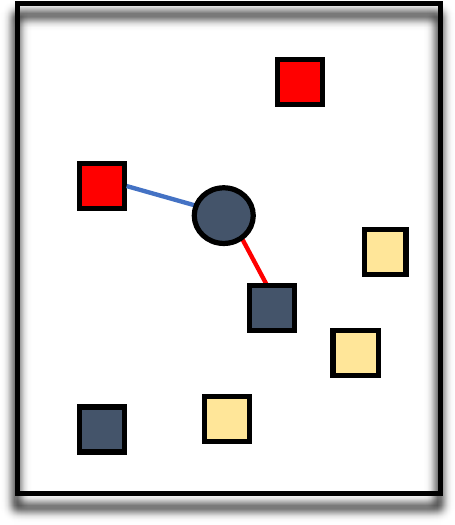}
         \caption{Triplet~\cite{Hoffer2015DeepML}}
         \label{fig:fig1a}
     \end{subfigure}
     \begin{subfigure}[b]{0.16\textwidth}
         \centering
         \includegraphics[width=\textwidth]{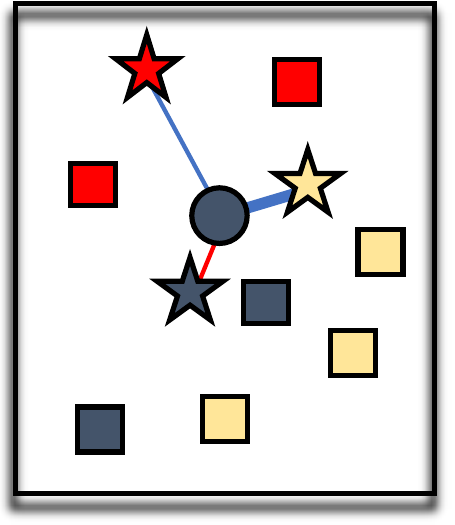}
         \caption{P-NCA~\cite{MovshovitzAttias2017NoFD}}
         \label{fig:fig1b}
     \end{subfigure}
     \begin{subfigure}[b]{0.16\textwidth}
         \centering
         \includegraphics[width=\textwidth]{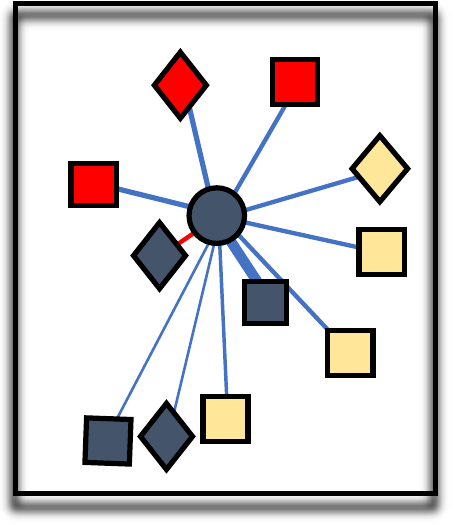}
         \caption{NT-Xent~\cite{Chen2020ASF}}
         \label{fig:fig1c}
     \end{subfigure}
     \begin{subfigure}[b]{0.16\textwidth}
         \centering
         \includegraphics[width=\textwidth]{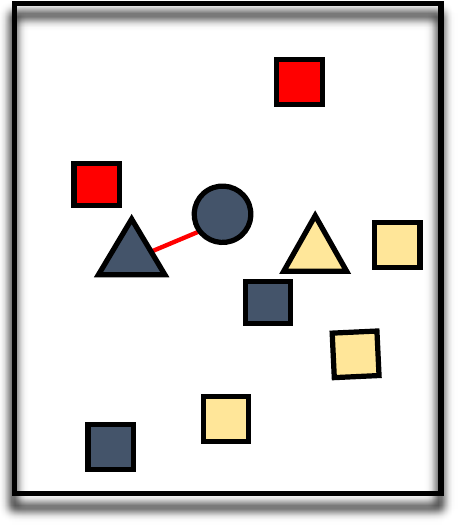}
         \caption{ACGAN~\cite{Odena2017ConditionalIS}}
         \label{fig:fig1d}
     \end{subfigure}
     \begin{subfigure}[b]{0.16\textwidth}
         \centering
         \includegraphics[width=\textwidth]{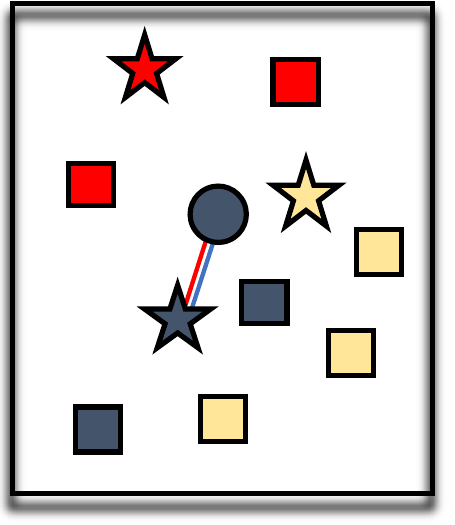}
         \caption{ProjGAN~\cite{Miyato2018cGANsWP}}
         \label{fig:fig1e}
     \end{subfigure}
     \begin{subfigure}[b]{0.16\textwidth}
         \centering
         \includegraphics[width=\textwidth]{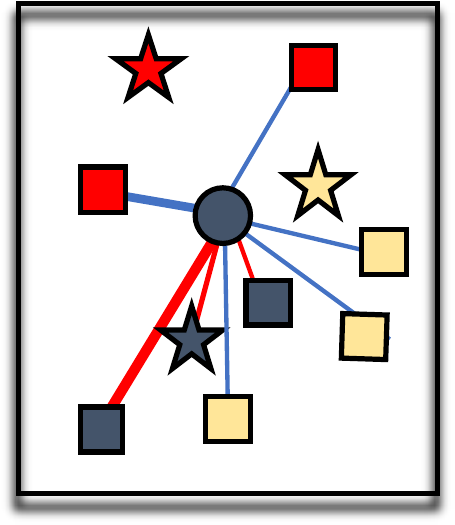}
         \caption{2C loss (Ours)}
         \label{fig:fig1f}
     \end{subfigure}
    \caption{Illustrative figures visualize metric learning losses (a,b,c) and conditional GANs (d,e,f). The color indicates the class label and the shape represents the role. (Square) the embedding of an image. (Diamond) the embedding of an augmented image. (Circle) a reference image embedding. Each loss is applied to the reference. (Star) the embedding of a class label. (Triangle) the one-hot encoding of a class label. The thicknesses of red and blue lines represent the strength of pull and push force, respectively. The loss function of ProjGAN lets the reference and the corresponding class embedding be close to each other when the reference is real, but it pushes far away otherwise. Compared to ACGAN and ProjGAN, 2C loss can consider both data-to-class and data-to-data relations between training examples.}
    \label{fig:metric_learning_losses}
\end{figure}

The goal of the discriminator in ACGAN is to classify the class of a given image and the sample's authenticity. Using data-to-class relations, i.e., information about which class a given data belongs to, the generator tries to generate fake images that can deceive the authenticity and are classified as the target labels. Since ACGAN uses a cross-entropy loss to classify the class of an image, we can regard the conditioning loss of ACGAN as a pair-based loss that can consider only data-to-class relations (see Fig.~\ref{fig:fig1d}). ProjGAN tries to maximize inner-product values between embeddings of real images and the corresponding target embeddings while minimizing the inner-product values when the images are fake. Since the discriminator of ProjGAN pushes and pulls the embeddings of images according to the authenticity and class information, we can think of the conditioning objective of ProjGAN as a pair-based loss that considers data-to-class relations~(see Fig.~\ref{fig:fig1e}). Unlike ACGAN, which looks at relations between a fixed one-hot vector and a sample, ProjGAN can consider more flexible relations using a learnable class embedding, namely Proxy.

\subsection{Conditional Contrastive Loss}
\label{sec:2c_loss}
To exploit data-to-data relations, we can adopt loss functions used in self-supervised~\cite{Chen2020ASF} learning or metric learning~\cite{Hoffer2015DeepML, Hadsell2006DimensionalityRB, Law2013QuadrupletWiseIS, Sohn2016ImprovedDM, Aziere2019EnsembleDM, Opitz2020DeepML}. In other words, our approach is to \emph{add a metric learning or self-supervised learning objective} in the \emph{discriminator} and \emph{generator} to explicitly control distances between embedded image features depending on the labels. Several metric learning losses, such as contrastive loss~\cite{Hadsell2006DimensionalityRB}, triplet loss~\cite{Hoffer2015DeepML}, quadruplet loss~\cite{Law2013QuadrupletWiseIS}, and N-pair loss~\cite{Sohn2016ImprovedDM} could be good candidates. However, it is known that 1) mining informative triplets or quadruplets requires higher training complexity, and 2) poor tuples make the training time longer. While the proxy-based losses~\cite{MovshovitzAttias2017NoFD, Aziere2019EnsembleDM, Opitz2020DeepML} relieves mining complexity using trainable class embedding vectors, such losses do not explicitly take data-to-data relations~\cite{kim2020proxy} into account.

Before introducing the proposed 2C loss, we bring NT-Xent loss~\cite{Chen2020ASF} to express our idea better. Let $\mX=\{\vx_1,..., \vx_m\}$, where $\vx \in \mathbb{R}^{W\times H}$ be a randomly sampled minibatch of training images and $\vy=\{y_1,..., y_m\}$, where $y \in \mathbb{R}$ be the collection of corresponding class labels. Then, we define a deep neural network encoder $S(\vx) \in \mathbb{R}^k$ and a projection layer that embeds onto a new unit hypersphere $h: \mathbb{R}^k \longrightarrow \mathbb{S}^d.$ Then, we can map the data space to the hypersphere using the composition of $l=h(S(\cdot))$. NT-Xent loss conducts random data augmentations $T$ on the training data $\mX$, and we denote it as $\mA = \{\vx_1, T(\vx_1), ..., \vx_m, T(\vx_m)\} = \{\va_1, ..., \va_{2m}\}$. Using the above, we can formulate NT-Xent loss as follows: 
\begin{align*}
    \ell(\va_i, \va_j;t) = 
    -\text{log}\bigg(\frac{\text{exp}(l(\va_{i})^\top l(\va_{j})/t)}
    {\sum_{k=1}^{2m}\1_\mathrm{k \neq i} \cdot \text{exp}(l(\va_{i})^\top l(\va_{k})/t)}\bigg),  
    \label{eq:eq6}\tag{6}
\end{align*}
where the scalar value $t$ is a temperature to control push and pull force.  In this work, we use the part of the discriminator network ($D_{\phi_1}$) before the fully connected layer as the encoder network ($S$) and use multi-layer perceptrons parameterized by $\varphi$ as the projection head ($h$). As a result, we can map the data space to the unit hypersphere using $l=h(D_{\phi{1}}(\cdot))$. 

However, Eq.~(\ref{eq:eq6}) requires proper data augmentations and can not consider data-to-class relations of training examples. To resolve these issues, we propose to use the \emph{embeddings of class labels} instead of using data augmentations. With a class embedding function $e(y): \mathbb{R} \longrightarrow \mathbb{R}^d$, Eq.~(\ref{eq:eq6}) can be formulated as follows:
\begin{align*}
    \ell(\vx_i, y_i;t) = -\text{log}\bigg(\frac{\text{exp}(l(\vx_i)^\top e(y_i)/t)} {\text{exp}(l(\vx_i)^\top e(y_i)/t) + \sum_{k=1}^{m}\1_\mathrm{k \neq i} \cdot \text{exp}(l(\vx_i)^\top l(\vx_k)/t)}\bigg).
    \label{eq:eq7}\tag{7}
\end{align*}
Eq.~(\ref{eq:eq7}) pulls a reference sample $\vx_i$ nearer to the class embedding $e(y_i)$ and pushes the others away. This scheme may push negative samples which have the same label as $y_i$. Therefore, we make an exception by adding cosine similarities of such negative samples in the numerator of Eq.~(\ref{eq:eq7}). The final loss function is as follows:
\begin{align*}
    \ell_{\text{2C}}(\vx_i, y_i;t) = -\text{log}\bigg(
    \frac{\text{exp}(l(\vx_i)^\top e(y_i)/t)+ \sum_{k=1}^m \1_\mathrm{y_k = y_i} \cdot \text{exp}(l(\vx_i)^\top l(\vx_k)/t)}
    {\text{exp}(l(\vx_i)^\top e(y_i)/t) + \sum_{k=1}^{m}\1_\mathrm{k \neq i} \cdot \text{exp}(l(\vx_i)^\top l(\vx_k)/t)}\bigg).
    \label{eq:eq8}\tag{8}
\end{align*}
Eq.~(\ref{eq:eq8}) is the proposed conditional contrastive loss (2C loss). Minimizing 2C loss will reduce distances between the embeddings of images with the same labels while maximizing the others. 2C loss explicitly considers the data-to-data relations $l(\vx_i)^\top l(\vx_k)$ and data-to-class relations $l(\vx_i)^\top e(y_i)$ without comprehensive mining of the training dataset and augmentations. 

\renewcommand{\algorithmicrequire}{\textbf{Input:}}
\renewcommand{\algorithmicprocedure}{\textbf{For}}
\renewcommand{\algorithmicensure}{\textbf{Output:}}
\begin{algorithm}[t!]
\caption{: Training ContraGAN}\label{alg:algorithm}
\begin{algorithmic}[1]
\Require{Learning rate: $\alpha_{1}, \alpha_{2}$. Adam hyperparameters~\cite{Kingma2015AdamAM}: $\beta_{1}, \beta_{2}$. Batch size: $m$. Temperature: $t$.}
\Statex \quad \; \# of discriminator iterations per single generator iteration: $n_{dis}$. Contrastive coefficient: $\lambda$.
\Statex \quad \; Parameters of the generator, the discriminator, and the projection layer: $(\theta, \phi, \varphi)$.
\Ensure{Optimized $(\theta, \phi, \varphi)$.}

\Statex
\State Initialize $(\theta, \phi, \varphi)$
\For{$\{1, ..., $ \# of training iterations$\}$} 
    \For{$\{1, ..., n_\mathrm{dis}\}$}
    \State Sample $\{(\vx_{i}, y_{i}^\mathrm{real})\}_{i=1}^{m} \sim p_\mathrm{real}(\rvx,\ry)$ \State Sample $\{\vz_{i}\}_{i=1}^{m} \sim p(\rz)$ and $\{y_{i}^\mathrm{fake}\}_{i=1}^{m} \sim p(\ry)$
    \State $\mathcal{L}_{C}^\mathrm{real} \longleftarrow \frac{1}{m}\sum_{i=1}^{m}\ell_\text{2C}(\vx_i, y_i^\mathrm{real}; t)$ 
    \Comment{Eq. (\ref{eq:eq8}) with real images.}
    \State $\mathcal{L}_{D} \longleftarrow \frac{1}{m} \sum_{i=1}^{m} \{D_{\phi}(G_{\theta}(\vz_i, y_i^\mathrm{fake}), y_i^\mathrm{fake}) - D_{\phi}(\vx_{i}, y_i^\mathrm{real})\} + \lambda \mathcal{L}_{C}^\mathrm{real}$
    \State $\phi, \varphi \longleftarrow \text{Adam}(\mathcal{L}_{D}, \alpha_{1}, \beta_1, \beta_2)$
    \EndFor
\State Sample $\{\vz_{i}\}_{i=1}^{m} \sim p(\rz)$ and $\{y_i^\mathrm{fake}\}_{i=1}^{m} \sim p(\ry)$
\State $\mathcal{L}_{C}^\mathrm{fake} \longleftarrow \frac{1} {m}\sum_{i=1}^{m}\ell_\text{2C}(G_{\theta}(\vz_i, y_i^\mathrm{fake}), y_i^\mathrm{fake}; t)$ \Comment{Eq.~(\ref{eq:eq8}) with fake images.}
\State $\mathcal{L}_{G} \longleftarrow -\frac{1}{m} \sum_{i=1}^{m} \{D_{\phi}(G_{\theta}(\vz_i, y_i^\mathrm{fake}), y_i^\mathrm{fake})\} + \lambda \mathcal{L}_{C}^\mathrm{fake}$
\State $\theta \longleftarrow \text{Adam}(\mathcal{L}_{G}, \alpha_{2}, \beta_1, \beta_2)$
\EndFor
\end{algorithmic}
\end{algorithm}

\subsection{Contrastive Generative Adversarial Networks}
\label{sec:contraGAN}
With proposed 2C loss, we describe the framework, called ContraGAN and introduce training procedures. Like the typical training procedures of GANs, ContraGAN has a discriminator training step and a generator training step that compute an adversarial loss. With this foundation, ContraGAN additionally calculates 2C loss using a set of real or fake images. Algorithm 1 shows the training procedures of the proposed ContraGAN. A notable aspect is that 2C loss is computed using $m$ real images in the discriminator training step and $m$ generated images in the generator training step.

In this manner, the discriminator updates itself by minimizing the distances between real image embeddings from the same class while maximizing it otherwise. By forcing the embeddings to relate via 2C loss, the discriminator can learn the fine-grained representations of real images. Similarly, the generator exploits the knowledge of the discriminator, such as intra-class characteristics and higher-order representations of the real images, to generate more realistic images. 

\subsection{Differences between 2C Loss and NT-Xent Loss}
NT-Xent loss~\cite{Chen2020ASF} is intended for unsupervised learning. It regards the augmented image as the positive sample to consider data-to-data relations between an original image and the augmented image. On the other hand, 2C loss utilizes weak supervision of label information. Therefore, compared with 2C loss, NT-Xent hardly gathers image embeddings of the same class, since there is no supervision from the label information. Besides, NT-Xent loss requires extra data augmentations and additional forward and backward propagations, which induce a few times of longer training time than the model with 2C loss.
\section{Experiments}
\subsection{Datasets and Evaluation Metric}
We perform conditional image generation experiments with CIFAR10~\cite{Krizhevsky2009LearningML}, Tiny ImageNet~\cite{Tiny}, and ImageNet~\cite{Deng2009ImageNetAL} datasets to compare the proposed approach with other approaches.

\textbf{CIFAR10} \cite{Krizhevsky2009LearningML} is a widely used benchmark dataset in many image generation works~\cite{Miyato2018SpectralNF, Brock2019LargeSG, Zhang2019ConsistencyRF, Zhao2020ImprovedCR, Wu2019LOGANLO, Miyato2018cGANsWP, Odena2017ConditionalIS}, and it contains $32\times 32$ pixels of color images for 10 different classes. The dataset consists of 60,000 images in total. It is divided into 50,000 images for training and 10,000 images for testing.

\textbf{Tiny ImageNet} \cite{Tiny} provides 120,000 color images in total. Image size is $64\times 64$ pixels, and the dataset consists of 200 categories. Each category has 600 images divided into 500, 50, and 50 samples for training, validation, and testing, respectively. Tiny ImageNet has 10$\times$ smaller number of images per class than CIFAR10, but it provides $20\times$ larger number of classes than CIFAR10. Compared to CIFAR10, Tiny ImageNet is selected to test a more challenging scenario -- the number of images per class is not plentiful, but the network needs to learn more categories. 

\textbf{ImageNet} \cite{Deng2009ImageNetAL} provides 1,281,167 and 50,000 color images for training and validation respectively, and the dataset consists of 1,000 categories. We crop each image using a square box whose length is the same as the shorter side of the image. The cropped images are rescaled to $128\times 128$ pixels.

\textbf{Frechet Inception Distance (FID)} is an evaluation metric used in all experiments in this paper. 
The FID proposed by Heusel~\etal~\cite{Heusel2017GANsTB} calculates Wasserstein-2 distance~\cite{Wasserstein1969MarkovPO} between the features obtained from real images and generated images using Inception-V3 network~\cite{Szegedy2016RethinkingTI}. Since FID is the distance between two distributions, \emph{lower} FID indicates \emph{better} results.

\subsection{Software}
There are various approaches that report strong FID scores, but it is not easy to reproduce the results because detailed specifications for training or ways to measure the results are not clearly stated. For instance, FID could be different depending on the choice of the reference images (training, validation, or testing datasets could be used). Besides, FID score of prior work is not consistent, depending on TensorFlow versions~\cite{tensorflow2015-whitepaper}. Therefore, we re-implement twelve state-of-the-art GANs~\cite{Radford2016UnsupervisedRL, Mao2017LeastSG, Lim2017GeometricG, Arjovsky2017WassersteinG, Gulrajani2017ImprovedTO, Kodali2018OnCA, Odena2017ConditionalIS, Miyato2018cGANsWP,  Miyato2018SpectralNF, Zhang2019SelfAttentionGA, Brock2019LargeSG, Zhang2019ConsistencyRF} to validate the proposed ContraGAN under the same condition. Our implementation carefully follows the principal concepts and the available specifications in the prior work. Experimental results show that the results from our implementation are superior to the numbers in the original papers~\cite{Miyato2018SpectralNF, Brock2019LargeSG} for the experiments using CIFAR10 dataset. We hope that our implementation would relieve efforts to compare various GAN pipelines.

\subsection{Experimental Setup}
To conduct a reliable assessment, all experiments that use CIFAR10 and Tiny ImageNet datasets are performed three times with different random seeds, and we report means and standard deviations of FIDs. Experiments using ImageNet are executed once, and we report the best performance during the training. We calculate FID using CIFAR10's test images and the same amount of generated images. For the experiments using Tiny ImageNet and ImageNet, we use the validation set with the same amount of generated images. All FID values reported in our paper are calculated using the PyTorch FID implementation~\cite{pytorchttur}.

Since spectral normalization~\cite{Miyato2018SpectralNF} has become an essential element in modern GAN training, we use hinge loss~\cite{Lim2017GeometricG} and apply spectral normalization on all architectures used in our experiments. We adopt modern architectures used in the papers: DCGAN~\cite{Radford2016UnsupervisedRL, Miyato2018SpectralNF}, ResGAN~\cite{He_2016_CVPR, Gulrajani2017ImprovedTO}, and BigGAN~\cite{Brock2019LargeSG}, and all details about the architectures are described in the supplement.

Since the conditioning strategy used in the generator of ACGAN differs from that of ProjGAN, we incorporate the generator's conditioning method in all experiments for a fair comparison. We use the conditional coloring transform~\cite{Dumoulin2017ALR, de_Vries, Miyato2018cGANsWP}, which is the method adopted by the original ProjGAN.

Before conducting the main experiments, we investigate performance changes according to the type of projection layer $h$ in~Eq.~(\ref{eq:eq8}) and batch size. Although Chen~\etal~\cite{Chen2020ASF} reports that contrastive learning can benefit from a higher-dimensional projection and a larger batch size, we found that the linear projection with batch size 64 for CIFAR10 and 1,024 for Tiny ImageNet performs the best. For the dimension of the projection layer, we select 512 for CIFAR10, 768 for Tiny ImageNet, and 1,024 for ImageNet experiments. We do a grid search to find a proper temperature ($t$) used in Eq.~\ref{eq:eq8} and experimentally found that the temperature of $1.0$ gives the best results. Detailed hyperparameter settings used in our experiments are described in the supplement.
\begin{table}[t!]
  \caption{Experiments on the effectiveness of 2C loss. Considering both data-to-data and data-to-class relations largely improves image generation results based on FID values. Mean$\pm$variance of FID is reported, and lower is better.}
  \label{table2}
  \vspace{3.0mm}
  \centering
  \resizebox{1.0\textwidth}{!}{
  \begin{tabular}{cccccc}
    \toprule
    Dataset & Uncond. GAN~\cite{Brock2019LargeSG}   & with P-NCA loss~\cite{MovshovitzAttias2017NoFD}  & with NT-Xent loss~\cite{Chen2020ASF} & with Eq.7 loss & with 2C loss (ContraGAN) \\
    \midrule
    CIFAR10 \cite{Krizhevsky2009LearningML} & 15.550$\pm$1.955 & 15.350$\pm$0.017  & 14.832$\pm$0.695 & 10.886$\pm$0.072 & \textbf{10.597}$\pm$\textbf{0.273} \\
    \midrule
    Tiny ImageNet \cite{Tiny} & 56.297$\pm$1.499 & 47.867$\pm$1.813 & 54.394$\pm$9.982 & 33.488$\pm$1.006 & \textbf{32.720}$\pm$\textbf{1.084} \\
    \bottomrule
  \end{tabular}
  }
\end{table}
\begin{table}[t!]
  \caption{Experiments using CIFAR10 and Tiny ImageNet datasets. Using three backbone architectures (DCGAN, ResGAN, and BigGAN), we test three approaches using different class conditioning models~(ACGAN, ProjGAN, and ContraGAN (ours)).} 
  \label{table1}
  \vspace{3.0mm}
  \centering
  \resizebox{0.9\textwidth}{!}{
  \begin{tabular}{ccccc}
    \toprule
    \multirow{2}{*}{Dataset} & \multirow{2}{*}{Backbone} & \multicolumn{3}{c}{Method for class information conditioning} \\
     & & ACGAN \cite{Odena2017ConditionalIS} & ProjGAN \cite{Miyato2018cGANsWP}  & ContraGAN (Ours)\\
    \midrule
    \multirow{3}{*}{CIFAR10~\cite{Krizhevsky2009LearningML}} & DCGAN~\cite{Radford2016UnsupervisedRL, Miyato2018SpectralNF} & 21.439$\pm$0.581&19.524$\pm$ 0.249&\textbf{18.788}$\pm$\textbf{0.571} \\
    & ResGAN~\cite{He_2016_CVPR, Gulrajani2017ImprovedTO} &11.588$\pm$0.093&\textbf{11.025}$\pm$ \textbf{0.177}&11.334$\pm$0.126 \\
    & BigGAN~\cite{Brock2019LargeSG} & 10.697$\pm$0.129&10.739$\pm$0.016&\textbf{10.597}$\pm$\textbf{0.273}\\
    \midrule
    \multirow{1}{*}{Tiny ImageNet~\cite{Tiny}} & BigGAN~\cite{Brock2019LargeSG}  & 88.628$\pm$5.523  & 37.563$\pm$4.390 & \textbf{32.720}$\pm$\textbf{1.084} \\
    \bottomrule
  \end{tabular}
  }
\end{table}
\vspace{-3mm}
\begin{table}[t!]
  \caption{Comparison with state-of-the-art GAN models. We mark `*' to FID values reported in the original papers~\cite{Miyato2018SpectralNF,Zhang2019SelfAttentionGA, Zhang2019ConsistencyRF}. The other FID values are obtained from our implementation. We conduct ImageNet~\cite{Deng2009ImageNetAL} experiments with a batch size of 256.} 
  \label{table3}
  \vspace{3.0mm}
  \centering
  \resizebox{1.0\textwidth}{!}{
  \begin{tabular}{cccccc}
    \toprule
    Dataset & SNResGAN \cite{Miyato2018SpectralNF}   & SAGAN \cite{Zhang2019SelfAttentionGA}  & BigGAN \cite{Brock2019LargeSG} & ContraGAN (Ours) & Improvement\\
    \midrule
    CIFAR10 \cite{Krizhevsky2009LearningML} & *17.5 & 17.127$\pm$0.220  & *14.73/10.739$\pm$0.016 & \textbf{10.597}$\pm$\textbf{0.273} & *+28.1\%/\textbf{+1.3}\%\\
    \midrule
    Tiny ImageNet \cite{Tiny} & 47.055$\pm$3.234 & 46.221$\pm$3.655 & 31.771$\pm$3.968 & \textbf{29.492}$\pm$\textbf{1.296} & \textbf{+7.2}\% \\
    \midrule
    ImageNet \cite{Deng2009ImageNetAL} & - & - & 21.072 & \textbf{19.443} & \textbf{+7.7}\%\\
    \bottomrule
  \end{tabular}
  }
\end{table}
\subsection{Evaluation Results}
\textbf{Effectiveness of 2C loss.} We compare 2C loss with P-NCA loss~\cite{MovshovitzAttias2017NoFD}, NT-Xent loss~\cite{Chen2020ASF}, and the objective function formulated in Eq.~\ref{eq:eq7}. P-NCA loss [24] does not explicitly look at data-to-data relations, and NT-Xent loss [25] (equivalent to Eq. 6) does not take data-to-class relations into account. Our 2C loss can take advantage of both losses. Compared with the Eq.~\ref{eq:eq7} loss, 2C loss considers cosine similarities of negative samples whose labels are the same as the positive image. The experimental results show that considering both \emph{data-to-class} and \emph{data-to-data} relations is effective and largely enhances image generation performance on CIFAR10 and Tiny ImageNet dataset. Besides, removing degenerating negative samples gives slightly better performances on CIFAR10 and Tiny ImageNet datasets.

\textbf{Comparison with other conditional GANs.} We compare ContraGAN with ACGAN~\cite{Odena2017ConditionalIS} and ProjGAN~\cite{Miyato2018cGANsWP}, since these approaches are representative models using class information conditioning. As shown in Table~\ref{table1}, our approach shows favorable performances in CIFAR10, but our approach exhibits larger variations. Examples of generated images is shown in Fig.~\ref{fig:results} (left). Experiments with Tiny ImageNet indicate that our ContraGAN is more effective when the target dataset is in the higher-dimensional space and has large inter-class variations.

\textbf{Comparison with state-of-the-art models.} We compare our method with SNResGAN~\cite{Miyato2018SpectralNF}, SAGAN~\cite{Zhang2019SelfAttentionGA}, and BigGAN~\cite{Brock2019LargeSG}. All of these approaches adopt ProjGAN~\cite{Miyato2018cGANsWP} for class information conditioning. We conduct all experiments on Tiny ImageNet and ImageNet datasets using the hyperparameter setting used in SAGAN~\cite{Zhang2019SelfAttentionGA}. We use our implementation of BigGAN for a fair comparison and report the best FID values during training.

If we consider the most recent work, CRGAN~\cite{Zhang2019ConsistencyRF}, ICRGAN~\cite{Zhao2020ImprovedCR}, and LOGAN~\cite{Wu2019LOGANLO} can generate more realistic images than BigGAN. 
Compared to such approaches, we show that our framework outperforms BigGAN by just adopting the proposed 2C loss. CRGAN and ICRGAN conduct explicit data augmentations during the training, which requires additional gradient calculations for backpropagation. Also, LOGAN needs one more feedforward and backpropagation processes for latent optimization. It takes twice as much time to train the model than standard GANs.


As a result, we identify how ContraGAN performs without data augmentations or latent optimization. Table \ref{table3} quantitatively shows that ContraGAN can synthesize images better than other state-of-the-art GAN models under the same conditions. Compared to BigGAN, ContraGAN improves the performances by 1.3\% on CIFAR10, 7.3\% on Tiny ImageNet, 7.7\% on ImageNet. If we use the reported number in BigGAN paper~\cite{Brock2019LargeSG}, the improvement is 29.9\% on CIFAR10.

\subsection{Training Stability of ContraGAN}
\begin{figure}[t]
    \centering
    \includegraphics[width=0.31\linewidth]{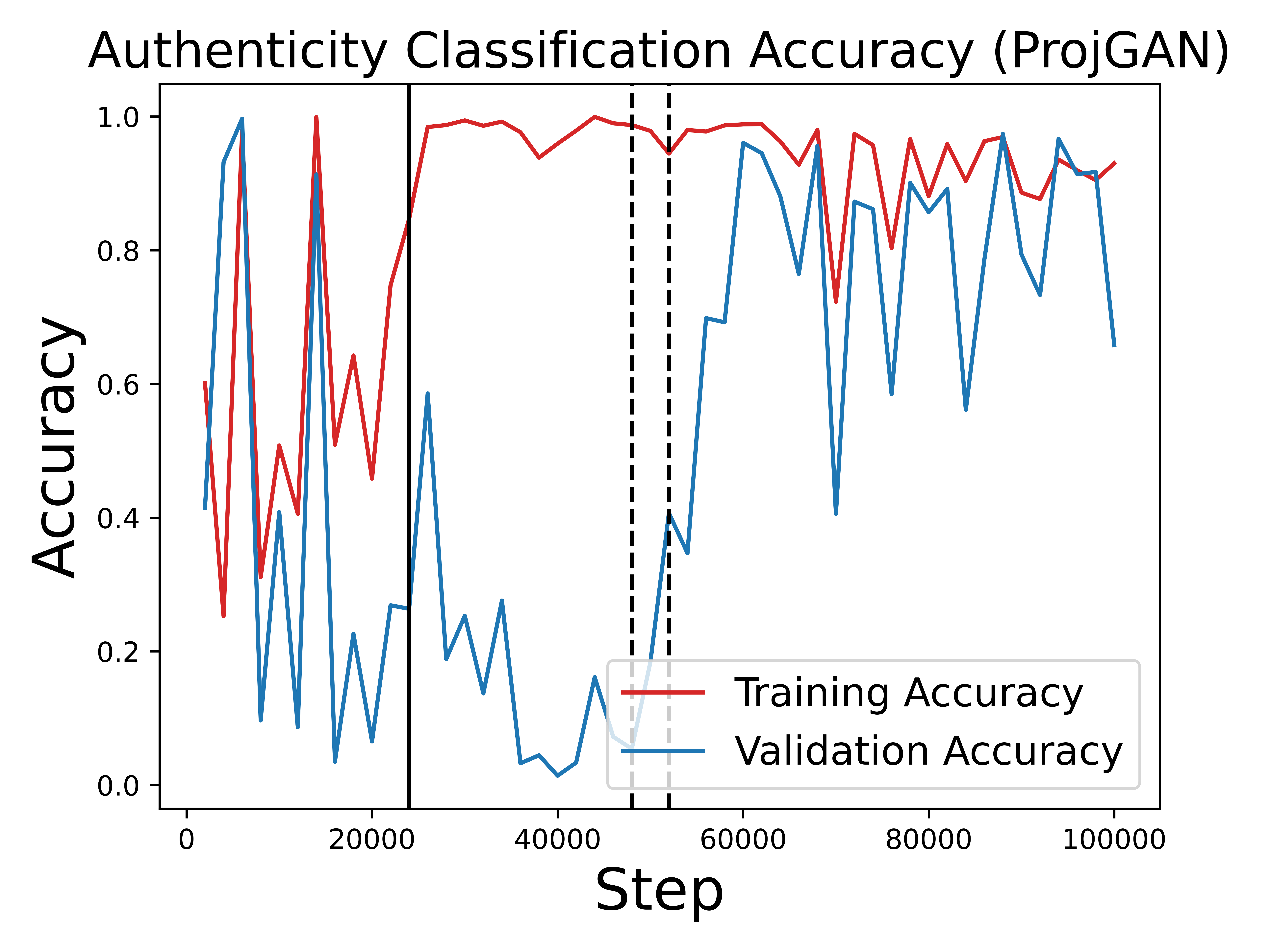}
    \includegraphics[width=0.31\linewidth]{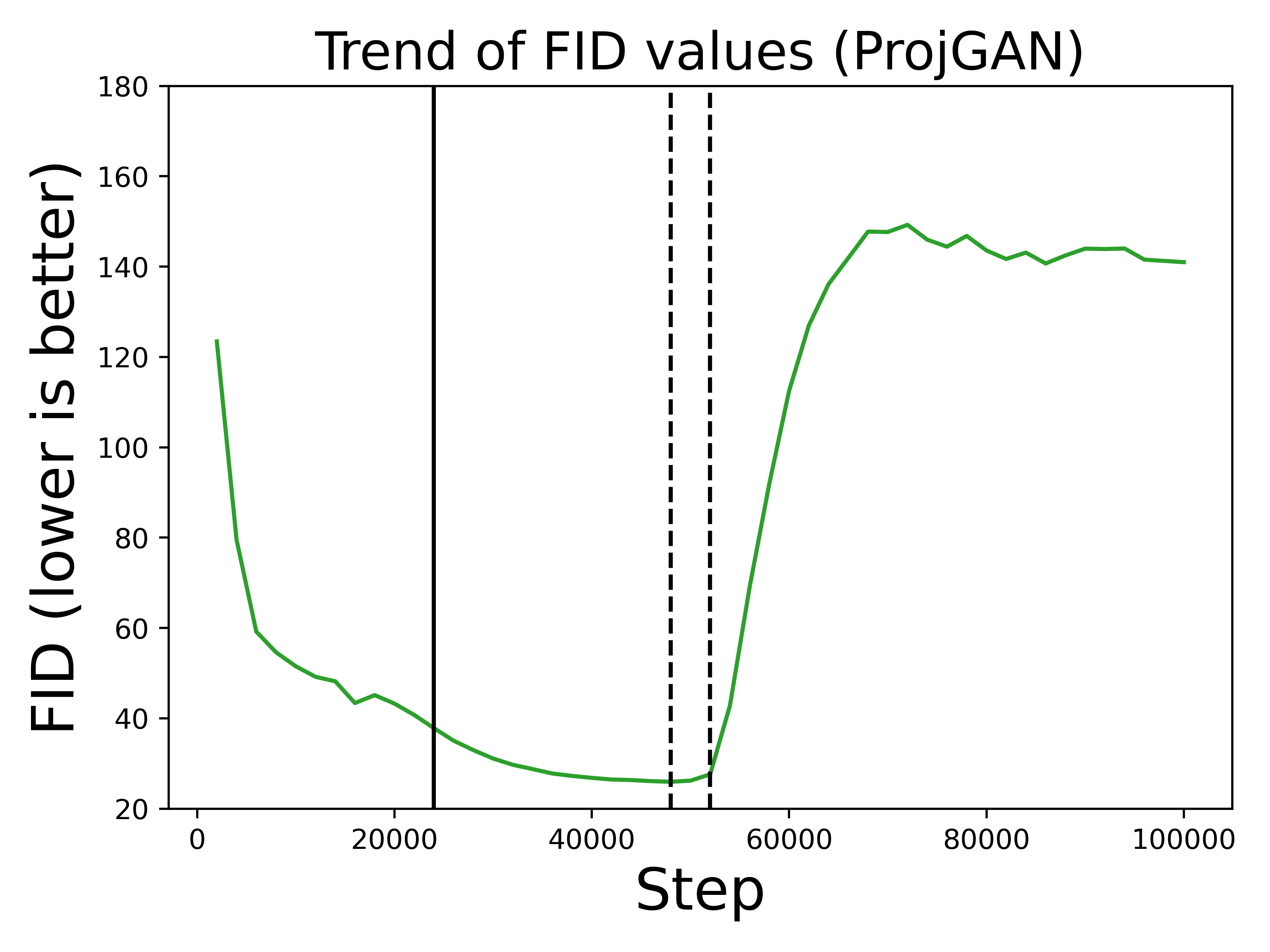}
    \includegraphics[width=0.31\linewidth]{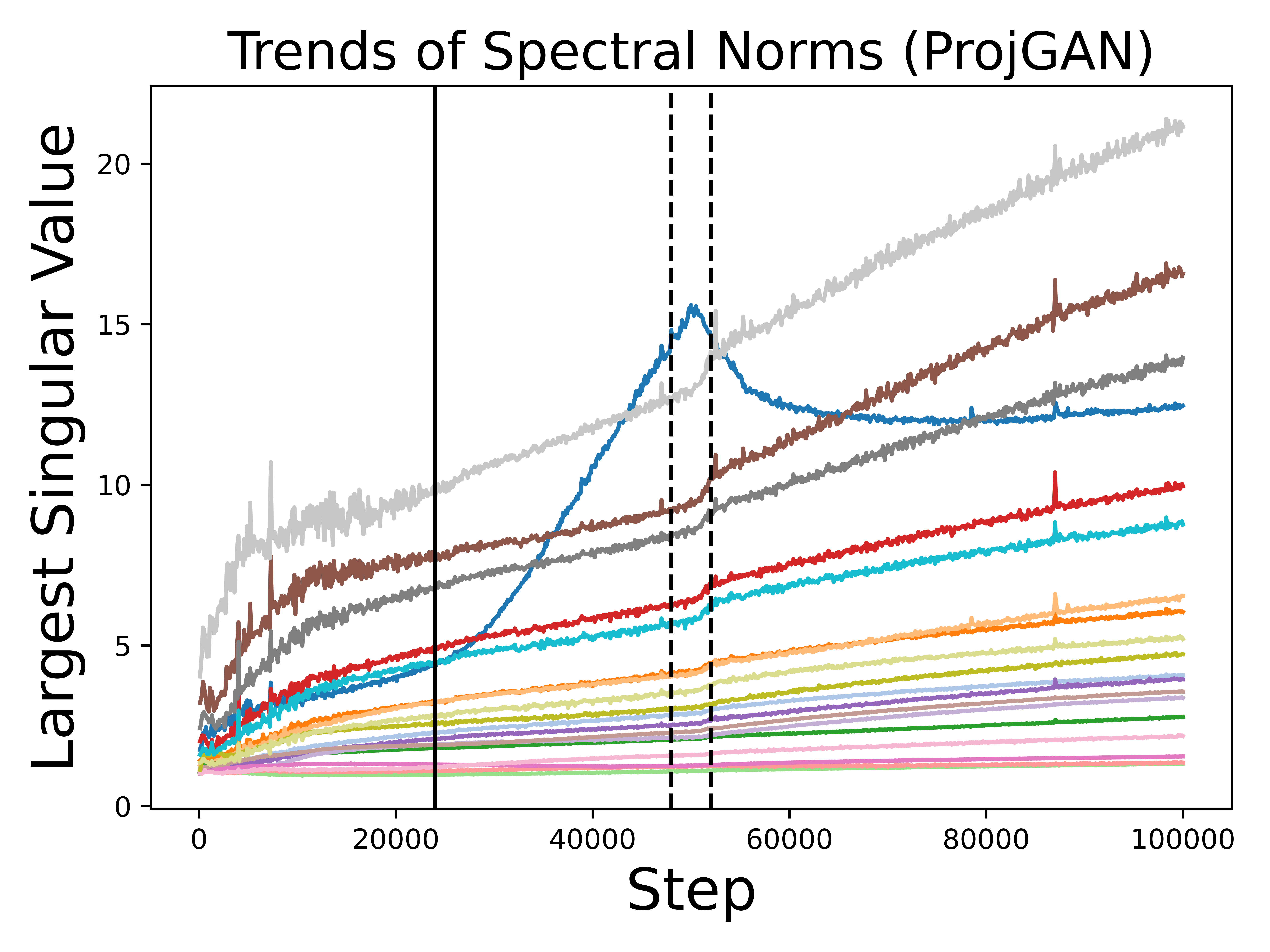}
    \includegraphics[width=0.31\linewidth]{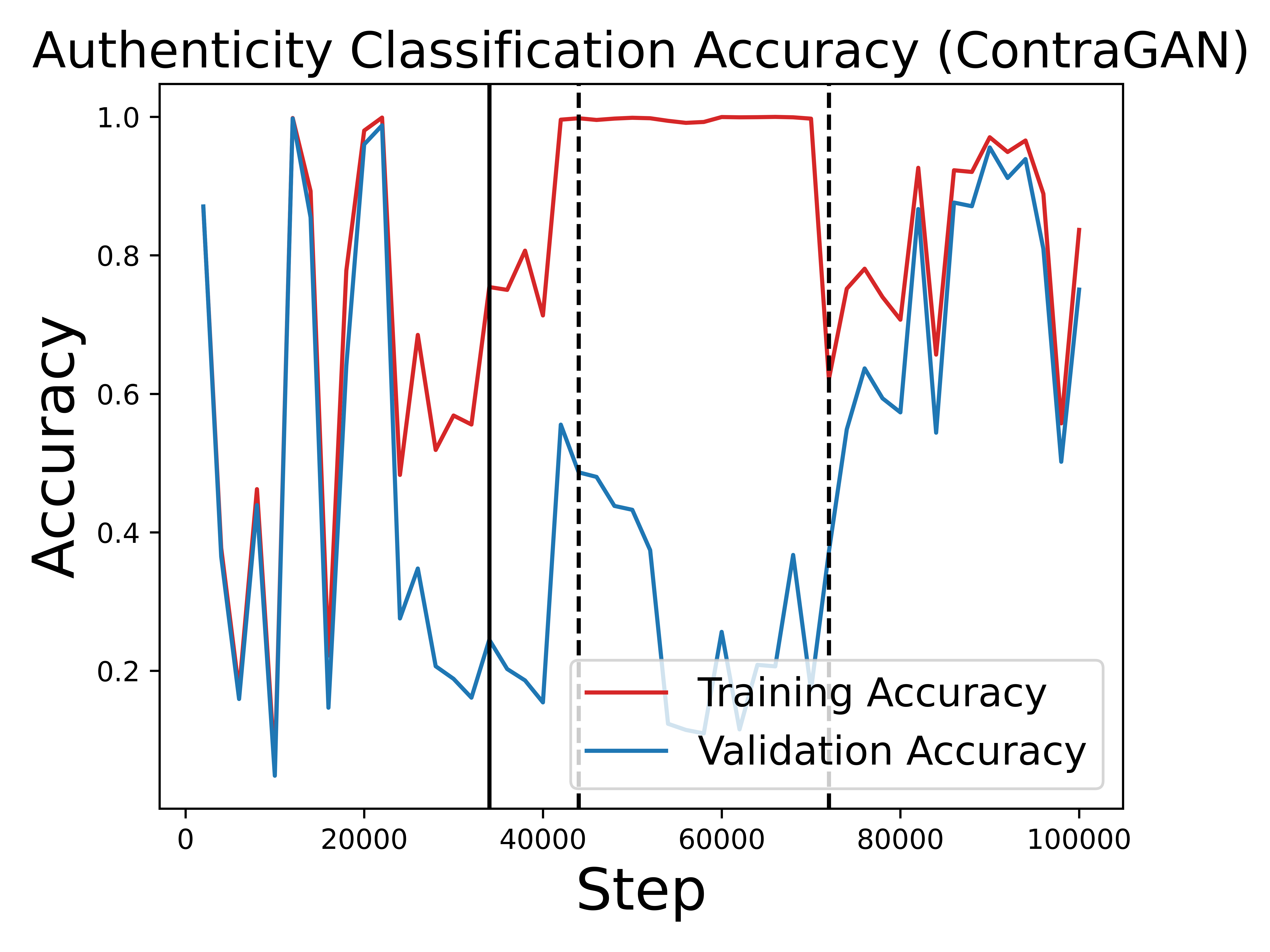}
    \includegraphics[width=0.31\linewidth]{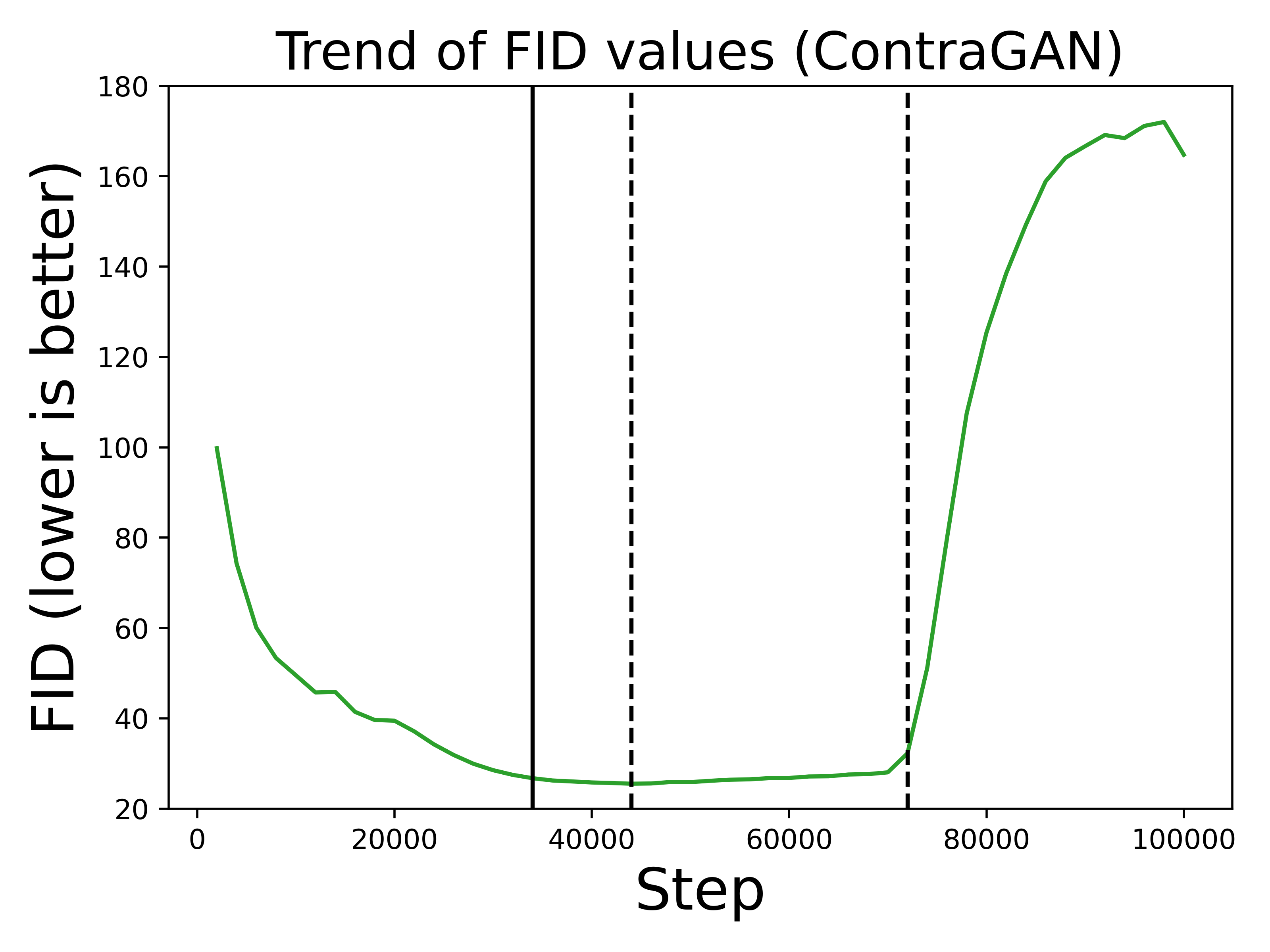}
    \includegraphics[width=0.31\linewidth]{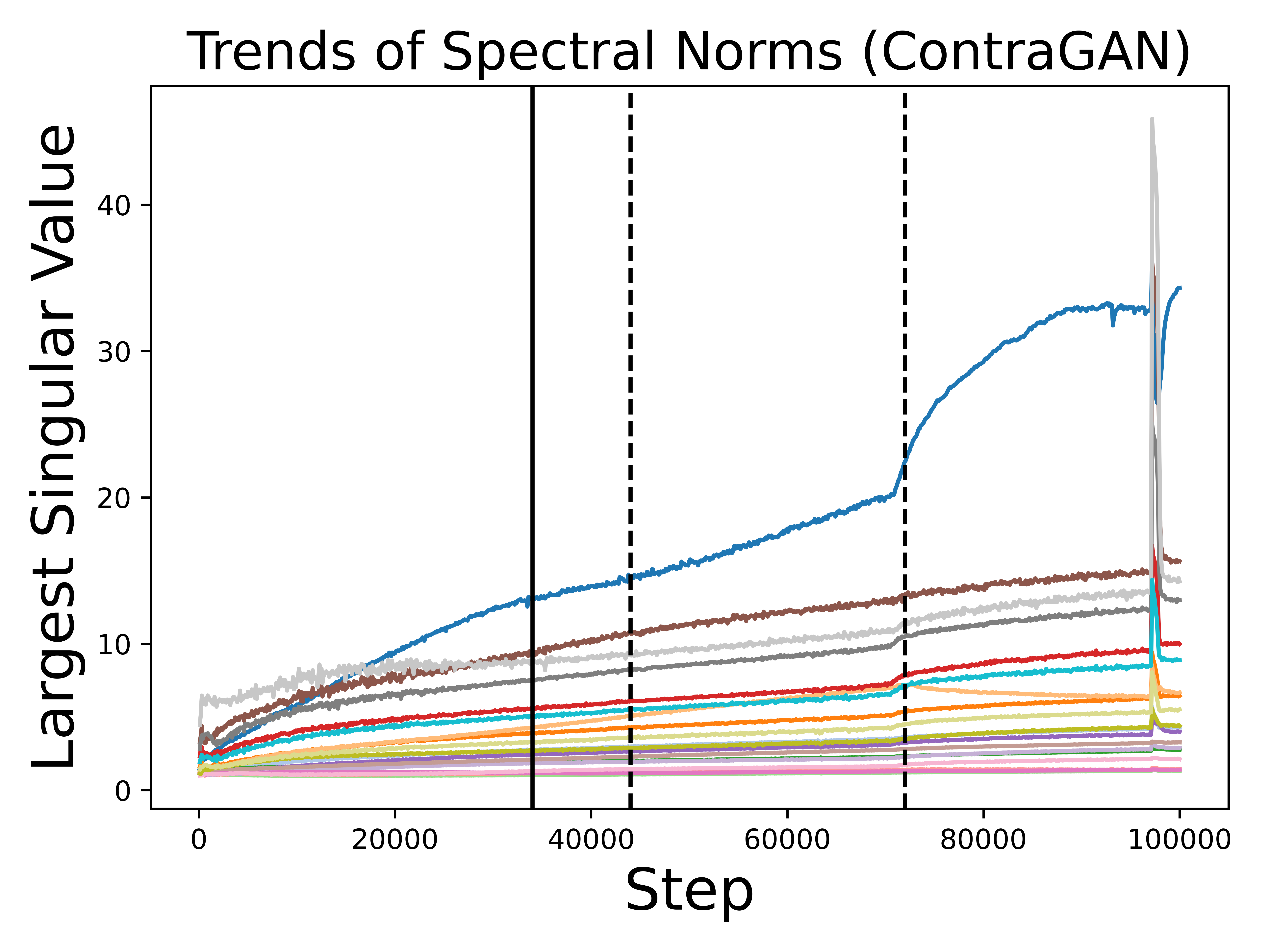}
    \caption{Authenticity classification accuracies on the training and validation datasets (left), trends of FID values (middle), and trends of the largest singular values of the discriminator's convolutional parameters (right). To specify the starting point where the difference between the training and validation accuracies is greater than 0.5, we use a solid black line. The first and second black dotted lines indicate when the performance is best and when training collapse occurs, respectively.}
    \label{fig:fig5}
\end{figure}
This section compares the training stability of ContraGAN and ProjGAN~\cite {Miyato2018cGANsWP} for the experiments using Tiny ImageNet. We compute the difference between the authenticity accuracies on the training and validation dataset. It is because the difference between training and validation performance is a good estimator for measuring the overfitting. Also, as Brock~\etal mentioned in his work~\cite{Brock2019LargeSG}, the sudden change in network parameters' largest singular values~(spectral norms) can indicate the collapse of adversarial training. Following this idea, we plot the trends of spectral norms of the discriminator's parameters to monitor the training collapse.

As shown in the first column of Fig.~\ref {fig:fig5}, ProjGAN shows the rapid increase of the accuracy difference, and ProjGAN reaches the collapse point earlier than ContraGAN. Moreover, the trend of FIDs and spectral analysis show that ContraGAN is more robust to training collapse. We speculate that ContraGAN is harder to reach undesirable status since ContraGAN jointly considers data-to-data and data-to-label relations. We discover that an increase in the accuracy on the validation dataset can indicate training collapse. 

\begin{figure}[ht]
    \centering
    \includegraphics[width=0.46\linewidth]{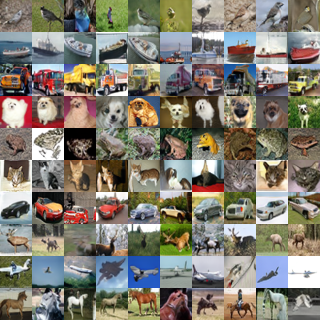}
    \includegraphics[width=0.46\linewidth]{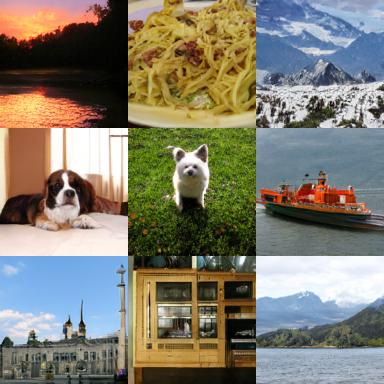}
    \caption{Examples of generated images using the proposed ContraGAN. (left) CIFAR10~\cite{Krizhevsky2009LearningML}, FID: 10.322, (right) ImageNet~\cite{Deng2009ImageNetAL}, FID: 19.443. In the case of ImageNet experiment, we select and plot well-generated images.}
    \label{fig:results}
\end{figure}

\vspace{-1mm}
\subsection{Ablation Study}
\label{sec:ablation}
We study how ContraGAN can be improved further with a larger batch size and data augmentations. We use ProjGAN with BigGAN architecture on Tiny ImageNet for this study. We use consistency regularization (CR)~\cite{Zhang2019ConsistencyRF} to identify that our ContraGAN can benefit from regularization that uses data augmentations. Also, to identify that 2C loss is not only computationally cheap but also effective to train GANs, we replace the class embeddings with augmented positive samples (APS). APS is widely used in the self-supervised contrastive learning community~\cite{Chen2020ASF, Khosla2020SupervisedCL}. Table~\ref{table4} shows the experiment settings, FID, and time per iteration. We indicate the number of parameters as Param. and denote three ablations -- (the 2C loss, augmented positive samples (APS), and consistency regularization (CR)) as Reg. 

\vspace{-1mm}
\textbf{Large batch size.} (A, C) and (E, H) show that ContraGAN can benefit from large batch size. 

\vspace{-1mm}
\textbf{Effect of the proposed 2C loss.} (A, E) and (C, H) show that the proposed 2C loss significantly reduces FID scores of the vanilla networks (A, C) by 21.6\% and 11.2\%, respectively. 

\vspace{-1mm}
\textbf{Comparison with APS.} From the experiments (E, F), we can see that the 2C loss performs better than 2C loss + APS, although the latter takes about 12.9\% more time. We speculate this is because each class embedding can become the representatives of the class, and it serves as the anchor that pulls corresponding images. Without the class embeddings, images in a minibatch are collected depending on a sampling state, and this may lead to training instability.

\vspace{-1mm}
\textbf{Comparison with CR.}
(A, E, G) and (C, H, I) show that vanilla + 2C loss + CR can reduce FIDs of either the results from vanilla networks (A, C) and vanilla + 2C loss (E, H). Note that the synergy is only observable if CR is used with 2C loss, and vanilla + 2C loss + CR beats vanilla + CR (B, D) with a large margin.
\begin{table}[h!]
  \caption{Ablation study on various batch sizes, losses, and regularizations. In Reg. row, we mark $-$ if an approach not applied and mark $\checkmark$ otherwise (in order of 2C loss, Augmented Positive Samples (APS), and Consistency Regularization~\cite{Zhang2019ConsistencyRF} (CR)). Please refer Sec.~\ref{sec:ablation} for the details.}
  \vspace{3mm}
  \label{table4}
  \centering
  \resizebox{1.0\textwidth}{!}{
  \begin{tabular}{cccccccccc}
    \toprule
    ID & (A) & (B) & (C) & (D) & (E) & (F) & (G) & (H) & (I) \\
    \midrule
    Batch & 256 & 256 & 1024 & 1024 & 256 & 256 & 256 & 1024 & 1024 \\ 
    Param. & 48.1 & 48.1 & 48.1 & 48.1 & 49.0 & 49.0 & 49.0 & 49.0 & 49.0\\
    Reg. & - - - & - - \checkmark & - - - & - - \checkmark & \checkmark - - & \checkmark\checkmark - & \checkmark - \checkmark & \checkmark - - & \checkmark - \checkmark\\
    FID & 40.981 & 36.434 & 34.090 & 38.231 & 32.094 & 33.151 & 28.631 & 30.286 & \textbf{27.018}\\
    Time & 0.901 & 1.093 & 3.586 & 4.448 & 0.967 & 1.110 & 1.121 & 3.807 & 4.611\\
    \bottomrule
  \end{tabular}
  }
   \vspace{-3mm}
\end{table}
\section{Conclusion}
In this paper, we formulate a conditional contrastive loss (2C loss) and present new Contrastive Generative Adversarial Networks (ContraGAN) for conditional image generation. Unlike previous conditioning losses, the proposed 2C loss considers not only data-to-class but also data-to-data relations between training examples. Under the same conditions, we demonstrate that ContraGAN outperforms state-of-the-art conditional GANs on Tiny ImageNet and ImageNet datasets. Also, we identify that ContraGAN helps to relieve the discriminator's overfitting problem and training collapse. As future work, we would like to theoretically and experimentally analyze how adversarial training collapses as the authenticity accuracy on the validation dataset increases. Also, we think that exploring advanced regularization techniques~\cite{Zhao2020ImprovedCR, Wu2019LOGANLO, zhao2020differentiable, karras2020training} is necessary to understand ContraGAN further.

\newpage
\begin{ack}
This work was supported by Institute of Information \& communications Technology Planning \& Evaluation (IITP) grant funded by the Korea government(MSIT) (No.2019-0-01906, Artificial Intelligence Graduate School Program(POSTECH)). The supercomputing resources for this work was partially supported by Grand Challenging Project of Supercomuting Bigdata Center, DGIST.
\end{ack}
\section*{Broader Impact}
We proposed a new conditional image generation model that can synthesize more realistic and diverse images. Our work can contribute to image-to-image translations~\cite{isola_i2i, 8237506}, generating realistic human faces~\cite{8953766, Karras2019AnalyzingAI, choi2020starganv2}, or any task that utilizes adversarial training. 

Since conditional GANs can expand to various image processing applications and can learn the representations of high-dimensional datasets, scientists can enhance the quality of astronomical images~\cite{Ledig2017PhotoRealisticSI, Schawinski2017GenerativeAN}, design complex architectured materials~\cite{Maoeaaz4169}, and efficiently search chemical space for developing materials~\cite{sampling_GAN}. 
We can do so many beneficial tasks with conditional GANs, but we should be concerned that conditional GANs can be used for deepfake techniques~\cite {Agarwal_2019_CVPR_Workshops}. Modern generative models can synthesize realistic images, making it more difficult to distinguish between real and fake. This can trigger sexual harassment~\cite{Xu2018FairGANFG}, fake news~\cite{NIPS2019_9106}, and even security issues of face recognition systems~\cite{deep_face_recog}.

To avoid improper use of conditional GANs, we need to be aware of generative models' strengths and weaknesses. Besides, it would be good to study the general characteristics of generated samples~\cite{wang2020cnn} and how we can distinguish fake images from unknown generative models \cite{8639163, Amerini_2019_ICCV, Celeb_DF_cvpr20}. 
{\small
    \bibliographystyle{unsrt}
    \bibliography{paper}

\begin{thebibliography}{10}

\bibitem{Goodfellow2014GenerativeAN}
Ian Goodfellow, Jean Pouget-Abadie, Mehdi Mirza, Bing Xu, David Warde-Farley,
  Sherjil Ozair, Aaron Courville, and Yoshua Bengio.
\newblock {Generative Adversarial Nets}.
\newblock In {\em Advances in Neural Information Processing Systems (NeurIPS)},
  pages 2672--2680, 2014.

\bibitem{Radford2016UnsupervisedRL}
Alec Radford, Luke Metz, and Soumith Chintala.
\newblock {Unsupervised Representation Learning with Deep Convolutional
  Generative Adversarial Networks}.
\newblock {\em arXiv preprint arXiv 1511.06434}, 2016.

\bibitem{Arjovsky2017WassersteinG}
Mart{\'i}n Arjovsky, Soumith Chintala, and L{\'e}on Bottou.
\newblock {Wasserstein GAN}.
\newblock {\em arXiv preprint arXiv 1701.07875}, 2017.

\bibitem{Miyato2018SpectralNF}
Takeru Miyato, Toshiki Kataoka, Masanori Koyama, and Yuichi Yoshida.
\newblock {Spectral Normalization for Generative Adversarial Networks}.
\newblock In {\em Proceedings of the International Conference on Learning
  Representations (ICLR)}, 2018.

\bibitem{Zhang2019SelfAttentionGA}
Han Zhang, Ian Goodfellow, Dimitris Metaxas, and Augustus Odena.
\newblock {Self-Attention Generative Adversarial Networks}.
\newblock In {\em Proceedings of the International Conference on Machine
  Learning (ICML)}, pages 7354--7363, 2019.

\bibitem{Brock2019LargeSG}
Andrew Brock, Jeff Donahue, and Karen Simonyan.
\newblock {Large Scale {GAN} Training for High Fidelity Natural Image
  Synthesis}.
\newblock In {\em Proceedings of the International Conference on Learning
  Representations (ICLR)}, 2019.

\bibitem{Zhang2019ConsistencyRF}
Han Zhang, Zizhao Zhang, Augustus Odena, and Honglak Lee.
\newblock {Consistency Regularization for Generative Adversarial Networks}.
\newblock In {\em Proceedings of the International Conference on Learning
  Representations (ICLR)}, 2020.

\bibitem{Zhao2020ImprovedCR}
Zhengli Zhao, Sameer Singh, Honglak Lee, Zizhao Zhang, Augustus Odena, and Han
  Zhang.
\newblock {Improved Consistency Regularization for GANs}.
\newblock {\em arXiv preprint arXiv 2002.04724}, 2020.

\bibitem{Wu2019LOGANLO}
Yan Wu, Jeff Donahue, David Balduzzi, Karen Simonyan, and Timothy~P. Lillicrap.
\newblock {LOGAN: Latent Optimisation for Generative Adversarial Networks}.
\newblock {\em arXiv preprint arXiv 1912.00953}, 2019.

\bibitem{Kodali2018OnCA}
Naveen Kodali, James Hays, Jacob~D. Abernethy, and Zsolt Kira.
\newblock {On Convergence and Stability of GANs}.
\newblock {\em arXiv preprint arXiv 1705.07215}, 2018.

\bibitem{Li2018OnTL}
Jerry Li, Aleksander Madry, John Peebles, and Ludwig Schmidt.
\newblock {On the Limitations of First-Order Approximation in GAN Dynamics}.
\newblock In {\em Proceedings of the International Conference on Machine
  Learning (ICML)}, 2018.

\bibitem{Nagarajan2017GradientDG}
Vaishnavh Nagarajan and J.~Zico Kolter.
\newblock {Gradient descent GAN optimization is locally stable}.
\newblock In {\em Advances in Neural Information Processing Systems (NeurIPS)},
  pages 5585--5595, 2017.

\bibitem{Mao2017LeastSG}
Xudong Mao, Qing Li, Haoran Xie, Raymond Y.~K. Lau, Zhixiang Wang, and
  Stephen~Paul Smolley.
\newblock {Least Squares Generative Adversarial Networks}.
\newblock In {\em Proceedings of the International Conference on Computer
  Vision (ICCV)}, pages 2813--2821, 2017.

\bibitem{Arjovsky2017TowardsPM}
Mart{\'i}n Arjovsky and L{\'e}on Bottou.
\newblock {Towards Principled Methods for Training Generative Adversarial
  Networks}.
\newblock In {\em Proceedings of the International Conference on Learning
  Representations (ICLR)}, 2017.

\bibitem{Lim2017GeometricG}
Jae~Hyun Lim and Jong~Chul Ye.
\newblock {Geometric GAN}.
\newblock {\em arXiv preprint arXiv 1705.02894}, 2017.

\bibitem{Gulrajani2017ImprovedTO}
Ishaan Gulrajani, Faruk Ahmed, Martin Arjovsky, Vincent Dumoulin, and Aaron~C
  Courville.
\newblock {Improved Training of Wasserstein GANs}.
\newblock In {\em Advances in Neural Information Processing Systems (NeurIPS)},
  pages 5767--5777, 2017.

\bibitem{Miyato2018cGANsWP}
Takeru Miyato and Masanori Koyama.
\newblock {c{GAN}s with Projection Discriminator}.
\newblock In {\em Proceedings of the International Conference on Learning
  Representations (ICLR)}, 2018.

\bibitem{Deng2009ImageNetAL}
Jia Deng, Wei Dong, Richard Socher, Li-Jia Li, Kai Li, and Fei-Fei Li.
\newblock {ImageNet: A large-scale hierarchical image database}.
\newblock In {\em Proceedings of the IEEE International Conference on Computer
  Vision and Pattern Recognition (CVPR)}, pages 248--255, 2009.

\bibitem{Odena2017ConditionalIS}
Augustus Odena, Christopher Olah, and Jonathon Shlens.
\newblock {Conditional Image Synthesis with Auxiliary Classifier {GAN}s}.
\newblock In {\em Proceedings of the International Conference on Machine
  Learning (ICML)}, pages 2642--2651, 2017.

\bibitem{Siarohin2019WhiteningAC}
Aliaksandr Siarohin, Enver Sangineto, and Nicu Sebe.
\newblock {Whitening and Coloring Batch Transform for GANs}.
\newblock In {\em Proceedings of the International Conference on Learning
  Representations (ICLR)}, 2019.

\bibitem{Brock2017NeuralPE}
Andrew Brock, Theodore Lim, James~M. Ritchie, and Nick Weston.
\newblock {Neural Photo Editing with Introspective Adversarial Networks}.
\newblock In {\em Proceedings of the International Conference on Learning
  Representations (ICLR)}, 2017.

\bibitem{zhao2020differentiable}
Shengyu Zhao, Zhijian Liu, Ji~Lin, Jun-Yan Zhu, and Song Han.
\newblock {Differentiable augmentation for data-efficient gan training}.
\newblock {\em arXiv preprint arXiv 2006.10738}, 2020.

\bibitem{karras2020training}
Tero Karras, Miika Aittala, Janne Hellsten, Samuli Laine, Jaakko Lehtinen, and
  Timo Aila.
\newblock {Training generative adversarial networks with limited data}.
\newblock {\em arXiv preprint arXiv 2006.06676}, 2020.

\bibitem{Krizhevsky2009LearningML}
Alex Krizhevsky.
\newblock {\em {Learning Multiple Layers of Features from Tiny Images}}.
\newblock PhD thesis, University of Toronto, 2012.

\bibitem{Tiny}
Johnson et~al.
\newblock {Tiny ImageNet Visual Recognition Challenge}.
\newblock \url{https://tiny-imagenet.herokuapp.com}.

\bibitem{He_2016_CVPR}
Kaiming He, Xiangyu Zhang, Shaoqing Ren, and Jian Sun.
\newblock {Deep Residual Learning for Image Recognition}.
\newblock In {\em Conference on Computer Vision and Pattern Recognition
  (CVPR)}, 2016.

\bibitem{ttur2010inline}
Martin Heusel, Hubert Ramsauer, Thomas Unterthiner, Bernhard Nessler, and Sepp
  Hochreiter.
\newblock {Two time-scale update rule for training GANs}.
\newblock \url{https://github.com/bioinf-jku/TTUR}, 2018.

\bibitem{Nash1951NONCOOPERATIVEG}
John Nash.
\newblock {Non-Cooperative Games}.
\newblock {\em Annals of mathematics}, pages 286--295, 1951.

\bibitem{Nowozin2016fGANTG}
Sebastian Nowozin, Botond Cseke, and Ryota Tomioka.
\newblock {f-{GAN}: Training Generative Neural Samplers using Variational
  Divergence Minimization}.
\newblock In {\em Advances in Neural Information Processing Systems (NeurIPS)},
  pages 271--279, 2016.

\bibitem{Sohn2015LearningSO}
Kihyuk Sohn, Honglak Lee, and Xinchen Yan.
\newblock Learning structured output representation using deep conditional
  generative models.
\newblock In {\em Advances in Neural Information Processing Systems (NeurIPS)},
  pages 3483--3491, 2015.

\bibitem{Mirza2014ConditionalGA}
Mehdi Mirza and Simon Osindero.
\newblock {Conditional Generative Adversarial Nets}.
\newblock {\em arXiv preprint arXiv 1411.1784}, 2014.

\bibitem{Hoffer2015DeepML}
Elad Hoffer and Nir Ailon.
\newblock {Deep Metric Learning Using Triplet Network}.
\newblock In {\em SIMBAD}, 2015.

\bibitem{MovshovitzAttias2017NoFD}
Yair Movshovitz-Attias, Alexander Toshev, Thomas~K. Leung, Sergey Ioffe, and
  Saurabh Singh.
\newblock {No Fuss Distance Metric Learning Using Proxies}.
\newblock In {\em Proceedings of the International Conference on Computer
  Vision (ICCV)}, pages 360--368, 2017.

\bibitem{Chen2020ASF}
Ting Chen, Simon Kornblith, Mohammad Norouzi, and Geoffrey~E. Hinton.
\newblock {A Simple Framework for Contrastive Learning of Visual
  Representations}.
\newblock {\em arXiv preprint arXiv 2002.05709}, 2020.

\bibitem{Hadsell2006DimensionalityRB}
Raia Hadsell, Sumit Chopra, and Yann LeCun.
\newblock {Dimensionality Reduction by Learning an Invariant Mapping}.
\newblock In {\em Proceedings of the IEEE International Conference on Computer
  Vision and Pattern Recognition (CVPR)}, pages 1735--1742, 2006.

\bibitem{Law2013QuadrupletWiseIS}
Marc~T. Law, Nicolas Thome, and Matthieu Cord.
\newblock {Quadruplet-Wise Image Similarity Learning}.
\newblock In {\em Proceedings of the International Conference on Computer
  Vision (ICCV)}, pages 249--256, 2013.

\bibitem{Sohn2016ImprovedDM}
Kihyuk Sohn.
\newblock {Improved Deep Metric Learning with Multi-class N-pair Loss
  Objective}.
\newblock In {\em Advances in Neural Information Processing Systems (NeurIPS)},
  pages 1857--1865, 2016.

\bibitem{Aziere2019EnsembleDM}
Nicolas Aziere and Sinisa Todorovic.
\newblock {Ensemble Deep Manifold Similarity Learning Using Hard Proxies}.
\newblock In {\em Proceedings of the IEEE International Conference on Computer
  Vision and Pattern Recognition (CVPR)}, pages 7291--7299, 2019.

\bibitem{Opitz2020DeepML}
Michael Opitz, Georg Waltner, Horst Possegger, and Horst Bischof.
\newblock {Deep Metric Learning with BIER: Boosting Independent Embeddings
  Robustly}.
\newblock {\em IEEE Transactions on Pattern Analysis and Machine Intelligence
  (TPAMI)}, 42:276--290, 2020.

\bibitem{kim2020proxy}
Sungyeon Kim, Dongwon Kim, Minsu Cho, and Suha Kwak.
\newblock {Proxy Anchor Loss for Deep Metric Learning}.
\newblock In {\em Proceedings of the IEEE International Conference on Computer
  Vision and Pattern Recognition (CVPR)}, 2020.

\bibitem{Kingma2015AdamAM}
Diederik~P. Kingma and Jimmy Ba.
\newblock {Adam: A Method for Stochastic Optimization}.
\newblock {\em arXiv preprint arXiv 1412.6980}, 2015.

\bibitem{Heusel2017GANsTB}
Martin Heusel, Hubert Ramsauer, Thomas Unterthiner, Bernhard Nessler, and Sepp
  Hochreiter.
\newblock {GANs Trained by a Two Time-Scale Update Rule Converge to a Local
  Nash Equilibrium}.
\newblock In {\em Advances in Neural Information Processing Systems (NeurIPS)},
  pages 6626--6637, 2017.

\bibitem{Wasserstein1969MarkovPO}
Leonid~Nisonovich Vaserstein.
\newblock {Markov processes over denumerable products of spaces, describing
  large systems of automata}.
\newblock {\em Problemy Peredachi Informatsii}, 5(3):64--72, 1969.

\bibitem{Szegedy2016RethinkingTI}
Christian Szegedy, Vincent Vanhoucke, Sergey Ioffe, Jon Shlens, and Zbigniew
  Wojna.
\newblock {Rethinking the Inception Architecture for Computer Vision}.
\newblock In {\em Proceedings of the IEEE International Conference on Computer
  Vision and Pattern Recognition (CVPR)}, pages 2818--2826, 2016.

\bibitem{tensorflow2015-whitepaper}
Mart\'{\i}n Abadi, Ashish Agarwal, Paul Barham, Eugene Brevdo, Zhifeng Chen,
  Craig Citro, Greg~S. Corrado, Andy Davis, Jeffrey Dean, Matthieu Devin,
  Sanjay Ghemawat, Ian Goodfellow, Andrew Harp, Geoffrey Irving, Michael Isard,
  Yangqing Jia, Rafal Jozefowicz, Lukasz Kaiser, Manjunath Kudlur, Josh
  Levenberg, Dandelion Man\'{e}, Rajat Monga, Sherry Moore, Derek Murray, Chris
  Olah, Mike Schuster, Jonathon Shlens, Benoit Steiner, Ilya Sutskever, Kunal
  Talwar, Paul Tucker, Vincent Vanhoucke, Vijay Vasudevan, Fernanda Vi\'{e}gas,
  Oriol Vinyals, Pete Warden, Martin Wattenberg, Martin Wicke, Yuan Yu, and
  Xiaoqiang Zheng.
\newblock {{TensorFlow}: Large-Scale Machine Learning on Heterogeneous
  Systems}, 2015.

\bibitem{pytorchttur}
{A Port of Fréchet Inception Distance (FID score) to PyTorch}.
\newblock \url{https://github.com/mseitzer/pytorch-fid}, 2018.

\bibitem{Dumoulin2017ALR}
Vincent Dumoulin, Jonathon Shlens, and Manjunath Kudlur.
\newblock {A Learned Representation For Artistic Style}.
\newblock In {\em Proceedings of the International Conference on Learning
  Representations (ICLR)}, 2017.

\bibitem{de_Vries}
Harm de~Vries, Florian Strub, Jeremie Mary, Hugo Larochelle, Olivier Pietquin,
  and Aaron~C Courville.
\newblock {Modulating early visual processing by language}.
\newblock In {\em Advances in Neural Information Processing Systems (NeurIPS)},
  pages 6594--6604, 2017.

\bibitem{Khosla2020SupervisedCL}
Prannay Khosla, Piotr Teterwak, Chen Wang, Aaron Sarna, Yonglong Tian, Phillip
  Isola, Aaron Maschinot, Ce~Liu, and Dilip Krishnan.
\newblock {Supervised Contrastive Learning}.
\newblock {\em arXiv preprint arXiv 2004.11362}, 2020.

\bibitem{isola_i2i}
T.~Zhou P.~Isola, J.~Zhu and A.~A. Efros.
\newblock {Image-to-Image Translation with Conditional Adversarial Networks}.
\newblock In {\em Proceedings of the IEEE International Conference on Computer
  Vision and Pattern Recognition (CVPR)}, pages 5967--5976, 2017.

\bibitem{8237506}
J.~{Zhu}, T.~{Park}, P.~{Isola}, and A.~A. {Efros}.
\newblock {Unpaired Image-to-Image Translation Using Cycle-Consistent
  Adversarial Networks}.
\newblock In {\em Proceedings of the International Conference on Computer
  Vision (ICCV)}, pages 2242--2251, 2017.

\bibitem{8953766}
T.~{Karras}, S.~{Laine}, and T.~{Aila}.
\newblock {A Style-Based Generator Architecture for Generative Adversarial
  Networks}.
\newblock In {\em Proceedings of the IEEE International Conference on Computer
  Vision and Pattern Recognition (CVPR)}, pages 4396--4405, 2019.

\bibitem{Karras2019AnalyzingAI}
Tero Karras, Samuli Laine, Miika Aittala, Janne Hellsten, Jaakko Lehtinen, and
  Timo Aila.
\newblock {Analyzing and Improving the Image Quality of StyleGAN}.
\newblock {\em arXiv preprint arXiv 1912.04958}, 2019.

\bibitem{choi2020starganv2}
Yunjey Choi, Youngjung Uh, Jaejun Yoo, and Jung-Woo Ha.
\newblock {StarGAN v2: Diverse Image Synthesis for Multiple Domains}.
\newblock In {\em Proceedings of the IEEE International Conference on Computer
  Vision and Pattern Recognition (CVPR)}, 2020.

\bibitem{Ledig2017PhotoRealisticSI}
Christian Ledig, Lucas Theis, Ferenc Husz{\'a}r, Jos{\'e}~Antonio Caballero,
  Andrew Aitken, Alykhan Tejani, Johannes Totz, Zehan Wang, and Wenzhe Shi.
\newblock {Photo-Realistic Single Image Super-Resolution Using a Generative
  Adversarial Network}.
\newblock In {\em Proceedings of the IEEE International Conference on Computer
  Vision and Pattern Recognition (CVPR)}, pages 105--114, 2017.

\bibitem{Schawinski2017GenerativeAN}
Kevin Schawinski, Ce~Zhang, Hantian Zhang, Lucas Fowler, and Gokula~Krishnan
  Santhanam.
\newblock {Generative Adversarial Networks recover features in astrophysical
  images of galaxies beyond the deconvolution limit}.
\newblock {\em arXiv preprint arXiv 1702.00403}, 2017.

\bibitem{Maoeaaz4169}
Yunwei Mao, Qi~He, and Xuanhe Zhao.
\newblock {Designing complex architectured materials with generative
  adversarial networks}.
\newblock {\em Science Advances}, 6(17), 2020.

\bibitem{sampling_GAN}
Yabo Dan, Yong Zhao, Xiang Li, Shaobo Li, Ming Hu, and Jianjun Hu.
\newblock {Generative adversarial networks (GAN) based efficient sampling of
  chemical composition space for inverse design of inorganic materials}.
\newblock {\em npj Computational Materials}, 6(1), 2020.

\bibitem{Agarwal_2019_CVPR_Workshops}
Shruti Agarwal, Hany Farid, Yuming Gu, Mingming He, Koki Nagano, and Hao Li.
\newblock {Protecting World Leaders Against Deep Fakes}.
\newblock In {\em The IEEE Conference on Computer Vision and Pattern
  Recognition (CVPR) Workshops}, 2019.

\bibitem{Xu2018FairGANFG}
Depeng Xu, Shuhan Yuan, Lu~Zhang, and Xintao Wu.
\newblock {FairGAN: Fairness-aware Generative Adversarial Networks}.
\newblock {\em International Conference on Big Data (Big Data)}, pages
  570--575, 2018.

\bibitem{NIPS2019_9106}
Rowan Zellers, Ari Holtzman, Hannah Rashkin, Yonatan Bisk, Ali Farhadi,
  Franziska Roesner, and Yejin Choi.
\newblock {Defending Against Neural Fake News}.
\newblock In {\em Advances in Neural Information Processing Systems (NeurIPS)},
  pages 9054--9065, 2019.

\bibitem{deep_face_recog}
Stephen Balaban.
\newblock {Deep learning and face recognition: the state of the art}.
\newblock In {\em Biometric and Surveillance Technology for Human and Activity
  Identification XII}, 2015.

\bibitem{wang2020cnn}
Sheng-Yu Wang, Oliver Wang, Richard Zhang, Andrew Owens, and Alexei~A Efros.
\newblock {CNN-generated images are surprisingly easy to spot... for now}.
\newblock In {\em Proceedings of the IEEE International Conference on Computer
  Vision and Pattern Recognition (CVPR)}, 2020.

\bibitem{8639163}
D.~{Güera} and E.~J. {Delp}.
\newblock {Deepfake Video Detection Using Recurrent Neural Networks}.
\newblock In {\em International Conference on Advanced Video and Signal Based
  Surveillance (AVSS)}, pages 1--6, 2018.

\bibitem{Amerini_2019_ICCV}
Irene Amerini, Leonardo Galteri, Roberto Caldelli, and Alberto Del~Bimbo.
\newblock {Deepfake Video Detection through Optical Flow Based CNN}.
\newblock In {\em International Conference on Computer Vision (ICCV)
  Workshops}, 2019.

\bibitem{Celeb_DF_cvpr20}
Yuezun Li, Pu~Sun, Honggang Qi, and Siwei Lyu.
\newblock {Celeb-DF: A Large-scale Challenging Dataset for DeepFake Forensics}.
\newblock In {\em Proceedings of the IEEE International Conference on Computer
  Vision and Pattern Recognition (CVPR)}, Seattle, WA, United States, 2020.

\bibitem{pmlr-v37-ioffe15}
Sergey Ioffe and Christian Szegedy.
\newblock {Batch Normalization: Accelerating Deep Network Training by Reducing
  Internal Covariate Shift}.
\newblock In {\em Proceedings of the International Conference on Machine
  Learning (ICML)}, pages 448--456, 2015.

\bibitem{Nair2010RectifiedLU}
Vinod Nair and Geoffrey~E. Hinton.
\newblock {Rectified Linear Units Improve Restricted Boltzmann Machines}.
\newblock In {\em Proceedings of the International Conference on Learning
  Representations (ICLR)}, 2010.

\bibitem{Xu2015EmpiricalEO}
Bing Xu, Naiyan Wang, Tianqi Chen, and Mu~Li.
\newblock {Empirical Evaluation of Rectified Activations in Convolutional
  Network}.
\newblock {\em arXiv preprint arXiv 1505.00853}, 2015.

\bibitem{Lin2014NetworkIN}
Min Lin, Qiang Chen, and Shuicheng Yan.
\newblock {Network In Network}.
\newblock {\em arXiv preprint arXiv 1312.4400}, 2014.

\bibitem{WardeFarley2017ImprovingGA}
David Warde-Farley and Yoshua Bengio.
\newblock {Improving Generative Adversarial Networks with Denoising Feature
  Matching}.
\newblock In {\em Proceedings of the International Conference on Learning
  Representations (ICLR)}, 2017.

\bibitem{Karras2018ProgressiveGO}
Tero Karras, Timo Aila, Samuli Laine, and Jaakko Lehtinen.
\newblock {Progressive Growing of GANs for Improved Quality, Stability, and
  Variation}.
\newblock {\em arXiv preprint arXiv 1710.10196}, 2018.

\bibitem{Mescheder2018ICML}
Lars Mescheder, Sebastian Nowozin, and Andreas Geiger.
\newblock {Which Training Methods for GANs do actually Converge?}
\newblock In {\em Proceedings of the International Conference on Machine
  Learning (ICML)}, 2018.

\bibitem{yazc2018the}
Yasin Yaz{\i}c{\i}, Chuan-Sheng Foo, Stefan Winkler, Kim-Hui Yap, Georgios
  Piliouras, and Vijay Chandrasekhar.
\newblock {The Unusual Effectiveness of Averaging in {GAN} Training}.
\newblock In {\em Proceedings of the International Conference on Learning
  Representations (ICLR)}, 2019.

\bibitem{NEURIPS2019_9015}
Adam Paszke, Sam Gross, Francisco Massa, Adam Lerer, James Bradbury, Gregory
  Chanan, Trevor Killeen, Zeming Lin, Natalia Gimelshein, Luca Antiga, Alban
  Desmaison, Andreas Kopf, Edward Yang, Zachary DeVito, Martin Raison, Alykhan
  Tejani, Sasank Chilamkurthy, Benoit Steiner, Lu~Fang, Junjie Bai, and Soumith
  Chintala.
\newblock {PyTorch: An Imperative Style, High-Performance Deep Learning
  Library}.
\newblock In {\em Advances in Neural Information Processing Systems (NeurIPS)},
  pages 8024--8035, 2019.

\end{thebibliography}
}

\clearpage
\appendix
\section*{Appendices}
\addcontentsline{toc}{section}{Appendices}
\renewcommand{\thesubsection}{\Alph{subsection}}
\renewcommand\thefigure{\thesection.\arabic{figure}}
\renewcommand{\thetable}{\Alph{section}\arabic{table}}
\setcounter{figure}{0}
\setcounter{table}{0}
\section{Network Architectures}
Since DCGAN~\cite{Radford2016UnsupervisedRL} showed astonishing image generation ability, several generator and discriminator architectures have been proposed to stabilize and enhance the generation quality. Representatively, Miyato~\etal~\cite{Miyato2018SpectralNF} have used a modified version of DCGAN~\cite{Radford2016UnsupervisedRL} and ResNet-style GAN~\cite{Gulrajani2017ImprovedTO} architectures with spectral normalization (We abbreviate it SNDCGAN and SNResGAN, respectively). Brock~\etal~\cite{Brock2019LargeSG} have expanded the capacity of SNResGAN with a shared embedding and skip connections from the noise vector (BigGAN). As a result, we tested the aforementioned frameworks to validate the proposed approach. To provide details of the main experiments in our paper, we introduce the network architectures in this section.

We start by defining some notations: $m$ is a batch size, \textsc{FC}(in\_features, out\_features) is a fully connected layer, \textsc{Conv}(in\_channels, out\_channels, kernel\_size, strides) is a convolutional layer, \textsc{Deconv}(in\_channels, out\_channels, kernel\_size, strides) is a deconvolutional layer, \textsc{BN} is a batch normalization~\cite{pmlr-v37-ioffe15}, \textsc{cBN} is a conditional batch normalization~\cite{Dumoulin2017ALR, de_Vries, Miyato2018cGANsWP}, \textsc{ReLU}, \textsc{LReLU}, and \textsc{Tanh} indicate ReLU~\cite{Nair2010RectifiedLU}, Leaky ReLU~\cite{Xu2015EmpiricalEO}, and hyperbolic tangent functions, respectively. \textsc{GBlock}(in channels, out channels, upsampling) is a generator block used in \cite{Gulrajani2017ImprovedTO, Miyato2018SpectralNF}, \textsc{BigGBlock}(in channels, out channels, upsampling, z split dim, shared dim) is a modified version of the \textsc{GBlock} used in~\cite{Brock2019LargeSG}, \textsc{DBlock}(in channels, out channels, downsampling) is a discriminator block used in~\cite{Brock2019LargeSG}, \textsc{Self-Attention} is a self-attention block used in~\cite{Zhang2019SelfAttentionGA}, \textsc{Normalize} is a normalize operation to project given embeddings onto a unit hypersphere, and \textsc{GSP} is a global sum pooling layer~\cite{Lin2014NetworkIN}. For more details about the \textsc{GBlock}, \textsc{BigGBlock}, \textsc{DBlock}, and the \textsc{Self-Attention} block, please refer to the papers \cite{Miyato2018SpectralNF, Zhang2019SelfAttentionGA, Brock2019LargeSG} or the code of our PyTorch implementation.

\begin{table}[h]
  \renewcommand\thetable{A\arabic{table}}
  \caption{Generator of SNDCGAN~\cite{Miyato2018SpectralNF} used for CIFAR10~\cite{Krizhevsky2009LearningML} image synthesis.}
  \label{SNDCGAN_G}
  \vspace{3.0mm}
  \centering
  \begin{tabular}{llrc}
    \toprule
    \textbf{Layer} & \textbf{Input}&\textbf{Output} & \textbf{Operation}\\
    \midrule
    Input Layer & (m, 128)&(m, 8192)&\textsc{FC}(128, 8192)\\
    \midrule
    Reshape Layer & (m, 8192)&(m, 4, 4, 512)&\textsc{Reshape}\\
    Hidden Layer & (m, 4, 4, 512)&(m, 8, 8, 256)&\textsc{Deconv}(512, 256, 4, 2),\textsc{cBN},\textsc{LReLU} \\
    Hidden Layer & (m, 8, 8, 256)&(m, 16, 16, 128)&\textsc{Deconv}(256, 128, 4, 2),\textsc{cBN},\textsc{LReLU} \\
    Hidden Layer & (m, 16, 16, 128)&(m, 32, 32, 64)&\textsc{Deconv}(128, 64, 4, 2),\textsc{cBN},\textsc{LReLU} \\
    Hidden Layer & (m, 32, 32, 64)&(m, 32, 32, 3)&\textsc{Conv}(64, 3, 3, 1) \\
    \midrule
    Output Layer & (m, 32, 32, 3)&(m, 32, 32, 3)&\textsc{Tanh} \\
    \bottomrule
  \end{tabular}
\end{table}
\begin{table}[h]
  \renewcommand\thetable{A\arabic{table}}
  \caption{Discriminator of SNDCGAN~\cite{Miyato2018SpectralNF} used for CIFAR10~\cite{Krizhevsky2009LearningML} image synthesis.}
  \label{SNDCGAN_D}
  \vspace{3.0mm}
  \centering
  \begin{tabular}{llrc}
    \toprule
    \textbf{Layer} & \textbf{Input} & \textbf{Output} & \textbf{Operation}\\
    \midrule
    Input Layer & (m, 32, 32, 3)  & (m, 32, 32, 64) & \textsc{Conv(3, 64, 3, 1)}, \textsc{LReLU}\\
    \midrule
    Hidden Layer & (m, 32, 32, 64)  & (m, 16, 16, 64) & \textsc{Conv(64, 64, 4, 2)}, \textsc{LReLU}\\
    Hidden Layer & (m, 16, 16, 64)  & (m, 16, 16, 128) & \textsc{Conv(64, 128, 3, 1)}, \textsc{LReLU}\\
    Hidden Layer & (m, 16, 16, 128)  & (m, 8, 8, 128) & \textsc{Conv(128, 128, 4, 2)}, \textsc{LReLU}\\
    Hidden Layer & (m, 8, 8, 128)  & (m, 8, 8, 256) & \textsc{Conv(128, 256, 3, 1)}, \textsc{LReLU}\\
    Hidden Layer & (m, 8, 8, 256)  & (m, 4, 4, 256) & \textsc{Conv(256, 256, 4, 2)}, \textsc{LReLU}\\
    Hidden Layer & (m, 4, 4, 256)  & (m, 4, 4, 512) & \textsc{Conv(256, 512, 3, 1)}, \textsc{LReLU}\\
    Hidden Layer & (m, 4, 4, 512)  & (m, 512) & \textsc{GSP}\\
    \midrule
    Output Layer & (m, 512)  & (m, 1) & \textsc{FC}(512, 1)\\
    \bottomrule
  \end{tabular}
\end{table}
\begin{table}[ht]
  \renewcommand\thetable{A\arabic{table}}
  \caption{Generator of SNResGAN~\cite{Miyato2018SpectralNF} used for CIFAR10~\cite{Krizhevsky2009LearningML} image synthesis.}
  \label{SNResGAN_G}
  \vspace{3.0mm}
  \centering
  \begin{tabular}{llrc}
    \toprule
    \textbf{Layer} & \textbf{Input} & \textbf{Output} & \textbf{Operation}\\
    \midrule
    Input Layer & (m, 128)  & (m, 4096) & \textsc{FC(128, 4096)}\\
    \midrule
    Reshape Layer & (m, 4096)  & (m, 4, 4, 256) & \textsc{Reshape}\\
    Hidden Layer & (m, 4, 4, 256) & (m, 8, 8, 256) & \textsc{GBlock}(256, 256, True) \\
    Hidden Layer & (m, 8, 8, 256) & (m, 16, 16, 256) & \textsc{GBlock}(256, 256, True) \\
    Hidden Layer & (m, 16, 16, 256) & (m, 32, 32, 256) & \textsc{GBlock}(256, 256, True) \\
    Hidden Layer & (m, 32, 32, 256) & (m, 32, 32, 3) & \textsc{BN}, \textsc{ReLU}, \textsc{Conv(256, 3, 3, 1)} \\
    \midrule
    Output Layer & (m, 32, 32, 3)  & (m, 32, 32, 3) & \textsc{Tanh} \\
    \bottomrule
    \vspace{20.0mm}
  \end{tabular}
\end{table}
\begin{table}[ht]
  \renewcommand\thetable{A\arabic{table}}
  \caption{Discriminator of SNResGAN~\cite{Miyato2018SpectralNF} used for CIFAR10~\cite{Krizhevsky2009LearningML} image synthesis.}
  \label{SNResGAN_D}
  \vspace{3.0mm}
  \centering
  \begin{tabular}{llrc}
    \toprule
    \textbf{Layer} & \textbf{Input} & \textbf{Output} & \textbf{Operation}\\
    \midrule
    Input Layer & (m, 32, 32, 3)  & (m, 16, 16, 128) & \textsc{DBlock}(3, 128, True)\\
    \midrule
    Hidden Layer & (m, 16, 16, 128)  & (m, 8, 8, 128) & \textsc{DBlock}(128, 128, True)\\
    Hidden Layer & (m, 8, 8, 128)  & (m, 8, 8, 128) & \textsc{DBlock}(128, 128, False)\\
    Hidden Layer & (m, 8, 8, 128)  & (m, 8, 8, 128) & \textsc{DBlock}(128, 128, False), \textsc{ReLU}\\
    Hidden Layer & (m, 8, 8, 128)  & (m, 128) & \textsc{GSP}\\
    \midrule
    Output Layer & (m, 128)  & (m, 1) & \textsc{FC(128, 1)}\\
    \bottomrule
    \vspace{20.0mm}
  \end{tabular}
\end{table}

\begin{table}[ht]
  \renewcommand\thetable{A\arabic{table}}
  \caption{Generator of BigGAN~\cite{Brock2019LargeSG} used for CIFAR10~\cite{Krizhevsky2009LearningML} image synthesis.}
  \label{BigGAN_G}
  \vspace{3.0mm}
  \centering
  \begin{tabular}{llrc}
    \toprule
    \textbf{Layer} & \textbf{Input} & \textbf{Output} & \textbf{Operation}\\
    \midrule
    Input Layer & (m, 20)  & (m, 6144) & \textsc{FC}(20, 6144)\\
    \midrule
    Reshape Layer & (m, 6144)  & (m, 4, 4, 384) & \textsc{Reshape}\\
    Hidden Layer & (m, 4, 4, 384) & (m, 8, 8, 384) & \textsc{BigGBlock}(384, 384, True, 20, 128) \\
    Hidden Layer & (m, 8, 8, 384) & (m, 16, 16, 384) & \textsc{BigGBlock}(384, 384, True, 20, 128) \\
    Hidden Layer & (m, 16, 16, 384) & (m, 16, 16, 384) & \textsc{Self-Attention} \\
    Hidden Layer & (m, 16, 16, 384) & (m, 32, 32, 384) & \textsc{BigGBlock}(384, 384, True, 20, 128) \\
    Hidden Layer & (m, 32, 32, 384) & (m, 32, 32, 3) & \textsc{BN}, \textsc{ReLU}, \textsc{Conv}(384, 3, 3, 1) \\
    \midrule
    Output Layer & (m, 32, 32, 3)  & (m, 32, 32, 3) & \textsc{Tanh} \\
    \bottomrule
  \end{tabular}
\end{table}
\begin{table}[ht]
  \renewcommand\thetable{A\arabic{table}}
  \caption{Discriminator of BigGAN~\cite{Brock2019LargeSG} used for CIFAR10~\cite{Krizhevsky2009LearningML} image synthesis.}
  \label{BigGAN_D}
  \vspace{3.0mm}
  \centering
  \begin{tabular}{llrc}
    \toprule
    \textbf{Layer} & \textbf{Input} & \textbf{Output} & \textbf{Operation}\\
    \midrule
    Input Layer & (m, 32, 32, 3)  & (m, 16, 16, 192) & \textsc{DBlock}(3, 192, True)\\
    \midrule
    Hidden Layer & (m, 16, 16, 192)  & (m, 16, 16, 192) & \textsc{Self-Attention}\\
    Hidden Layer & (m, 16, 16, 192)  & (m, 8, 8, 192) & \textsc{DBlock}(192, 192, True)\\
    Hidden Layer & (m, 8, 8, 192)  & (m, 8, 8, 192) & \textsc{DBlock}(192, 192, False)\\
    Hidden Layer & (m, 8, 8, 192)  & (m, 8, 8, 192) & \textsc{DBlock}(192, 192, False)\\
    Hidden Layer & (m, 8, 8, 192)  & (m, 192) &\textsc{ReLU}, \textsc{GSP}\\
    \midrule
    Output Layer & (m, 192)  & (m, 1) & \textsc{FC}(192, 1)\\
    \bottomrule
  \end{tabular}
\end{table}
\begin{table}[ht]
  \renewcommand\thetable{A\arabic{table}}
  \caption{Generator of BigGAN~\cite{Brock2019LargeSG} for Tiny ImageNet~\cite{Tiny} image synthesis.}
  \label{BigGAN64_G}
  \vspace{3.0mm}
  \centering
  \begin{tabular}{llrc}
    \toprule
    \textbf{Layer} & \textbf{Input} & \textbf{Output} & \textbf{Operation}\\
    \midrule
    Input Layer & (m,20)  & (m,20480) & \textsc{FC(20, 20480)}\\
    \midrule
    Reshape Layer & (m,20480)&(m,4,4,1280) & \textsc{Reshape}\\
    Hidden Layer & (m,4, 4, 1280)&(m,8, 8, 640) & \textsc{BigGBlock}(1280, 640, True, 20, 128) \\
    Hidden Layer & (m,8, 8, 640)&(m,16, 16, 320) & \textsc{BigGBlock}(640, 320, True, 20, 128) \\
    Hidden Layer & (m,16, 16, 320)&(m,32, 32, 160) & \textsc{BigGBlock}(320, 160, True, 20, 128) \\
    Hidden Layer & (m,32, 32, 160)&(m,32, 32, 160) & \textsc{Self-Attention} \\
    Hidden Layer & (m,32, 32, 160)&(m,64, 64, 80) & \textsc{BigGBlock}(160, 80, True, 20, 128) \\
    Hidden Layer & (m,64, 64, 80)&(m,64, 64, 3) & \textsc{BN}, \textsc{ReLU}, \textsc{Conv(80,3, 3, 1)} \\
    \midrule
    Output Layer & (m,32, 32, 3)&(m,32, 32, 3) & \textsc{Tanh} \\
    \bottomrule
  \end{tabular}
\end{table}
\begin{table}[ht]
  \renewcommand\thetable{A\arabic{table}}
  \caption{Discriminator of BigGAN~\cite{Brock2019LargeSG} for Tiny ImageNet~\cite{Tiny} image synthesis.}
  \label{BigGAN64_D}
  \vspace{3.0mm}
  \centering
  \begin{tabular}{llrc}
    \toprule
    \textbf{Layer} & \textbf{Input} & \textbf{Output} & \textbf{Operation}\\
    \midrule
    Input Layer & (m, 64, 64, 3)  & (m, 32, 32, 80) & \textsc{DBlock}(3, 80, True)\\
    \midrule
    Hidden Layer & (m, 32, 32, 80)  & (m, 32, 32, 80) & \textsc{Self-Attention}\\
    Hidden Layer & (m, 32, 32, 80)  & (m, 16, 16, 160) & \textsc{DBlock}(80, 160, True)\\
    Hidden Layer & (m, 16, 16, 160)  & (m, 8, 8, 320) & \textsc{DBlock}(160, 320, True)\\
    Hidden Layer & (m, 8, 8, 320)  & (m, 4, 4, 640) & \textsc{DBlock}(320, 640, True)\\
    Hidden Layer & (m, 4, 4, 640)  & (m, 4, 4, 1280) & \textsc{DBlock}(640, 1280, False)\\
    Hidden Layer & (m, 4, 4, 1280)  & (m, 1280) & \textsc{ReLU}, \textsc{GSP}\\
    \midrule
    Output Layer & (m, 1280)  & (m, 1) & \textsc{FC(1280, 1)}\\
    \bottomrule
  \end{tabular}
\end{table}
\clearpage
\begin{table}[ht]
  \renewcommand\thetable{A\arabic{table}}
  \caption{Generator of BigGAN~\cite{Brock2019LargeSG} for ImageNet~\cite{Deng2009ImageNetAL} image synthesis.}
  \label{BigGAN128_G}
  \vspace{3.0mm}
  \centering
  \begin{tabular}{llrc}
    \toprule
    \textbf{Layer} & \textbf{Input} & \textbf{Output} & \textbf{Operation}\\
    \midrule
    Input Layer & (m,20)  & (m,24576) & \textsc{FC(20, 24576)}\\
    \midrule
    Reshape Layer & (m,24576)&(m,4,4,1536) & \textsc{Reshape}\\
    Hidden Layer & (m,4,4,1536)&(m,8,8,1536) & \textsc{BigGBlock}(1536, 1536, True, 20, 128) \\
    Hidden Layer & (m,8,8,1536)&(m,16,16,768) & \textsc{BigGBlock}(1536, 768, True, 20, 128) \\
    Hidden Layer & (m,16,16,768)&(m,32,32,384) & \textsc{BigGBlock}(768, 384, True, 20, 128) \\
    Hidden Layer & (m,32,32,384)&(m,64,64,192) & \textsc{BigGBlock}(384, 192, True, 20, 128) \\
    Hidden Layer & (m,64,64,192)&(m,64,64,192) & \textsc{Self-Attention} \\
    Hidden Layer & (m,64,64,192)&(m,128,128,96) & \textsc{BigGBlock}(192, 96, True, 20, 128) \\
    Hidden Layer & (m,128,128,96)&(m,128,128,3) & \textsc{BN}, \textsc{ReLU}, \textsc{Conv(96, 3, 3, 1)} \\
    \midrule
    Output Layer & (m,128,128,3)&(m,128,128,3) & \textsc{Tanh} \\
    \bottomrule
  \end{tabular}
\end{table}
\begin{table}[ht]
  \renewcommand\thetable{A\arabic{table}}
  \caption{Discriminator of BigGAN~\cite{Brock2019LargeSG} for ImageNet~\cite{Deng2009ImageNetAL} image synthesis.}
  \label{BigGAN128_D}
  \vspace{3.0mm}
  \centering
  \begin{tabular}{llrc}
    \toprule
    \textbf{Layer} & \textbf{Input} & \textbf{Output} & \textbf{Operation}\\
    \midrule
    Input Layer & (m, 128, 128, 3)  & (m, 64, 64, 96) & \textsc{DBlock}(3, 96, True)\\
    \midrule
    Hidden Layer & (m, 64, 64, 96)  & (m, 64, 64, 96) & \textsc{Self-Attention}\\
    Hidden Layer & (m, 64, 64, 96)  & (m, 32, 32, 192) & \textsc{DBlock}(96, 192, True)\\
    Hidden Layer & (m, 32, 32, 192)  & (m, 16, 16, 384) & \textsc{DBlock}(192, 384, True)\\
    Hidden Layer & (m, 16, 16, 384)  & (m, 8, 8, 768) & \textsc{DBlock}(384, 768, True)\\
    Hidden Layer & (m, 8, 8, 768)  & (m, 4, 4, 1536) & \textsc{DBlock}(768, 1536, True)\\
    Hidden Layer & (m, 4, 4, 1536)  & (m, 4, 4, 1536) & \textsc{DBlock}(1536, 1536, False)\\
    Hidden Layer & (m, 4, 4, 1536)  & (m, 1536) & \textsc{ReLU}, \textsc{GSP}\\
    \midrule
    Output Layer & (m, 1536)  & (m, 1) & \textsc{FC(1536, 1)}\\
    \bottomrule
  \end{tabular}
\end{table}
\section{Hyperparameter Setup}
\begin{figure}[t!]
    \renewcommand\thefigure{A\arabic{figure}}
    \centering
    \includegraphics[scale=.40]{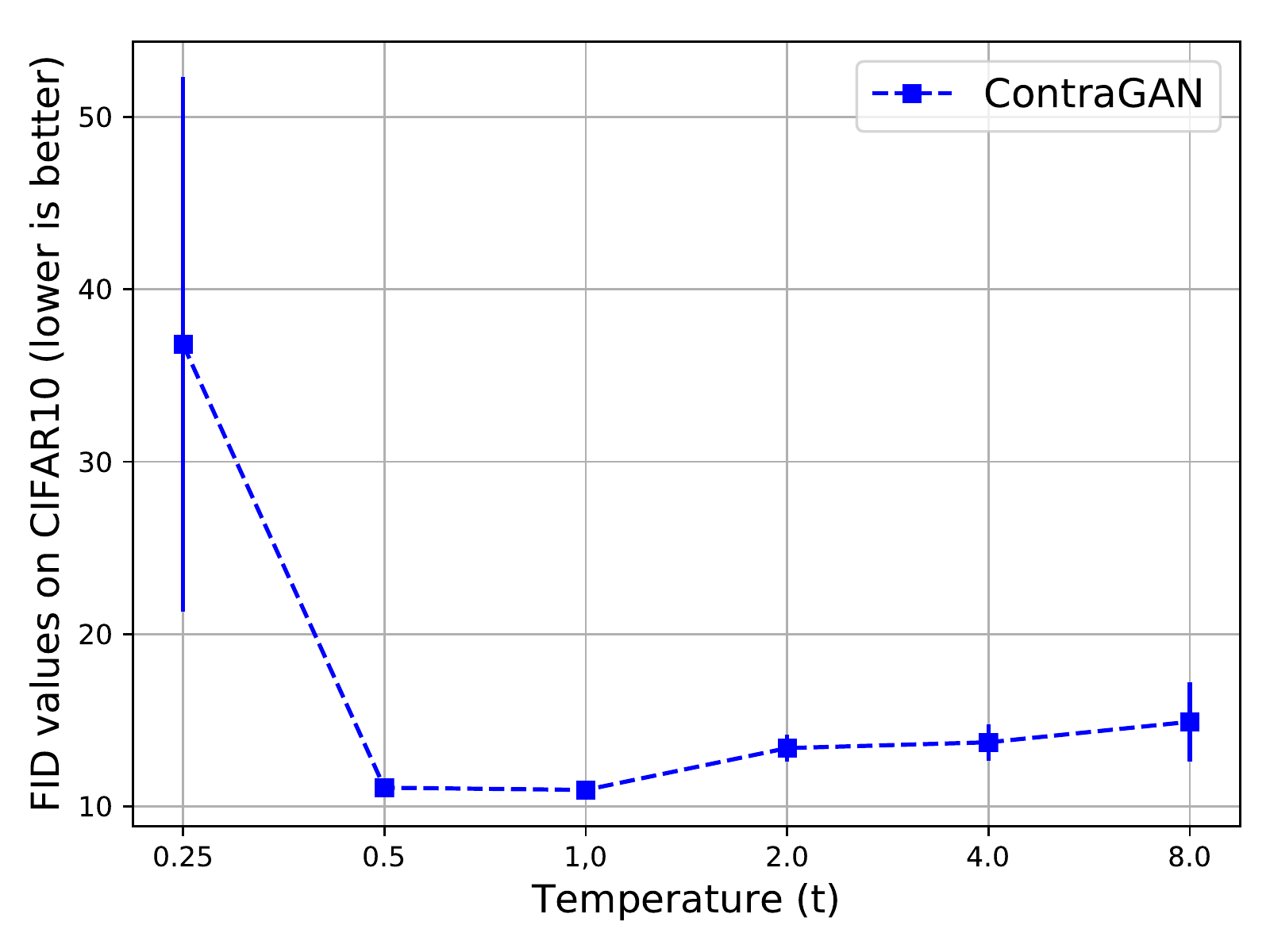}
    \caption{Change of FID values as the temperature increases. Experiments are executed three times, and the means and standard deviations are represented by the blue dots and solid lines, respectively.} 
    \label{FigureA2}
\end{figure}

\begin{table}[h]
  \renewcommand\thetable{A\arabic{table}}
  \caption{Hyperparameter values used for experiments. Settings (B, C, E) and (F) are the settings used in~\cite{WardeFarley2017ImprovingGA, Radford2016UnsupervisedRL, Zhang2019ConsistencyRF} and ~\cite{Zhang2019SelfAttentionGA}, respectively. we conduct experiments with CIFAR10~\cite{Krizhevsky2009LearningML} using the settings (A, B, C, D, E) and with Tiny ImageNet~\cite{Tiny} and ImageNet~\cite{Deng2009ImageNetAL} using the setting (F).}
  \label{tableA9}
  \vspace{3.0mm}
  \centering
  \begin{tabular}{cccccccc}
    \toprule
    Setting & $\alpha_1$ &  $\alpha_2$ & $\beta_1$  & $\beta_2$ & $n_{dis}$\\
    \midrule
    A & 0.0001 &  0.0001 & 0.5  & 0.999 & 2 \\
    B & 0.0001 &  0.0001 & 0.5  & 0.999 & 1 \\
    C & 0.0002 &  0.0002 & 0.5  & 0.999 & 1 \\
    D & 0.0002 &  0.0002 & 0.5  & 0.999 & 2 \\
    E & 0.0002 &  0.0002 & 0.5  & 0.999 & 5 \\
    F & 0.0004 &  0.0001 & 0.0  & 0.999 & 1 \\
    \bottomrule
  \end{tabular}
\end{table}
Choosing a proper hyperparameter setup is crucial to train GANs. In this paper, we conduct experiments using six settings with Adam optimizer~\cite{Kingma2015AdamAM}. $\alpha_1$ and $\alpha_2$ are the learning rates of the discriminator and generator. $\beta_1$ and $\beta_2$ are the hyperparameters of Adam optimizer to control exponential decay rates of moving averages. $n_{dis}$ is the number of discriminator iterations per single generator iteration. For the contrastive coefficient $\lambda$ (see Algorithm 1), the value is fixed at 1.0 for a fair comparison with~\cite{Odena2017ConditionalIS, Miyato2018cGANsWP}. In all experiments, we use the temperature $t=1.0$. Experiments over temperature are displayed in Fig.~\ref{FigureA2}. Besides, we apply moving average update of the generator's weights used in~\cite{Karras2018ProgressiveGO, Mescheder2018ICML, yazc2018the} after 20,000 generator iterations with the decay rate of 0.9999.
The settings (B, C, E) are known to give satisfactory performances on CIFAR10~\cite{Krizhevsky2009LearningML} in previous papers~\cite{WardeFarley2017ImprovingGA, Radford2016UnsupervisedRL, Zhang2019ConsistencyRF}. Since Heusel~\etal~\cite{Heusel2017GANsTB} and Zhang~\etal~\cite{Zhang2019SelfAttentionGA} have shown that two time-scale update (TTUR) can converge to a stationary local Nash equilibrium~\cite{Nash1951NONCOOPERATIVEG}, we adopt the hyperparameter setup used in~\cite{Zhang2019SelfAttentionGA} (setting F) to generate realistic images on Tiny ImageNet~\cite{Tiny} and ImageNet~\cite{Deng2009ImageNetAL} datasets.

\textbf{Experimental setup used for Table 1 in the main paper}: Experiments on CIFAR10 dataset are conducted three times with different random seeds using the setting~(E) with the batch size of 64 until 80k generator updates. Experiments on Tiny ImageNet dataset are performed three times until 100k generator updates using the setting~(F) with the batch size of 256 and BigGAN architecture (see Table \ref{BigGAN64_G} and Table \ref{BigGAN64_D}). 

\textbf{Experimental setup used for Table 2 in the main paper}: Experiments on CIFAR10 dataset are performed three times with different random seeds using the settings~(A, B, C, D, E) with the batch size of 64. We stop training GANs with SNDCGAN, SNResGAN, and BigGAN architectures after 200k, 100k, and 80k generator updates, respectively. Also, we report performances of the hyperparameter settings that showed the lowest FID values by mean. Experiments on Tiny ImageNet dataset are conducted three times until 100k generator updates using the setting~(F) with the batch size of 256 and BigGAN architecture (see Table \ref{BigGAN64_G} and Table \ref{BigGAN64_D}). The hyperparameter settings: C, D, E, show the best performance in SNDCGAN~\cite{Miyato2018SpectralNF}, SNResGAN~\cite{Miyato2018SpectralNF}, and BigGAN~\cite{Brock2019LargeSG}, respectively. We reason that as the model's capacity increases, training GANs becomes more difficult; thus, it requires more discriminator updates.  Moreover, we experimentally identify that 
updating discriminator more times does not always produce better performance, but it might be related to the model capacity.

\textbf{Experimental setup used for Table 3 in the main paper}: FID values on CIFAR10 dataset are reported using the setting (E) with the batch size of 64. The experiments on the Tiny ImageNet are conducted using the setting (F) with the batch size of 1024. Experiments on ImageNet dataset are executed once until 250k generator updates using the setting~(F) with the batch size of 256 and BigGAN architecture (see Table \ref{BigGAN128_G} and Table \ref{BigGAN128_D}). All other settings not noticed here are the same as the experimental setup for Table 2 above.

\textbf{Experimental setup used for Table 4 in the main paper}: All ablation results are reported using the setting (F), and models with consistency regularization~(CR)~\cite{Zhang2019ConsistencyRF} are trained with the coefficient of 10.0. We use an Intel(R) Xeon(R) Silver 4114 CPU, four NVIDIA Geforce RTX 2080 Ti GPUs, and PyTorch DataParallel library to measure time per iteration. All other settings not noticed here are the same as the experimental settings used for Table 2.
\section{Nonlinear Projection and Batch Size}
\begin{figure}[ht]
     \renewcommand\thefigure{A\arabic{figure}}
     \centering
     \begin{subfigure}[b]{0.48\textwidth}
         \centering
         \includegraphics[width=\textwidth]{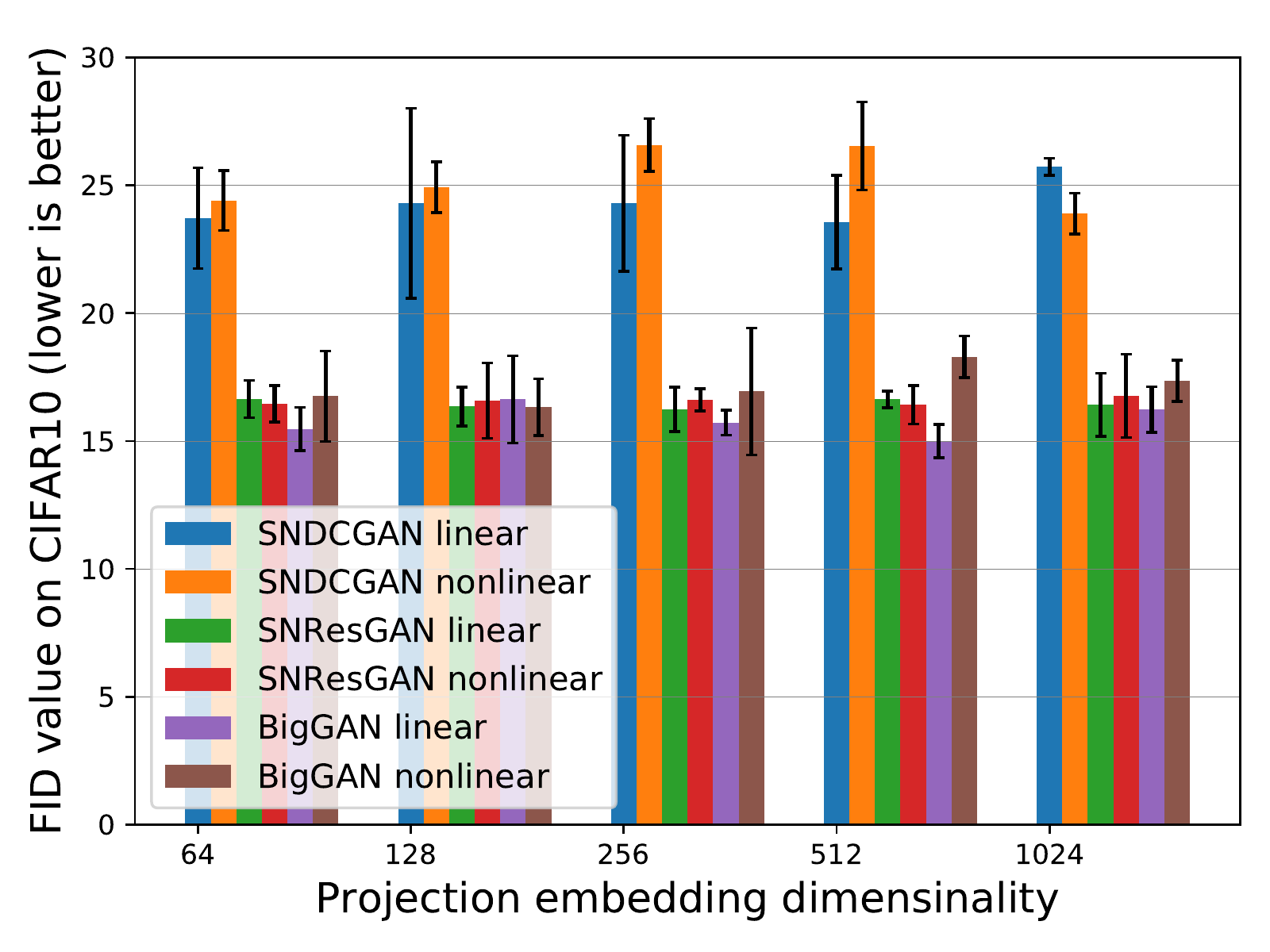}
         \caption{}
         \label{FigureA1a}
     \end{subfigure}
     \hfill
     \begin{subfigure}[b]{0.48\textwidth}
         \centering
         \includegraphics[width=\textwidth]{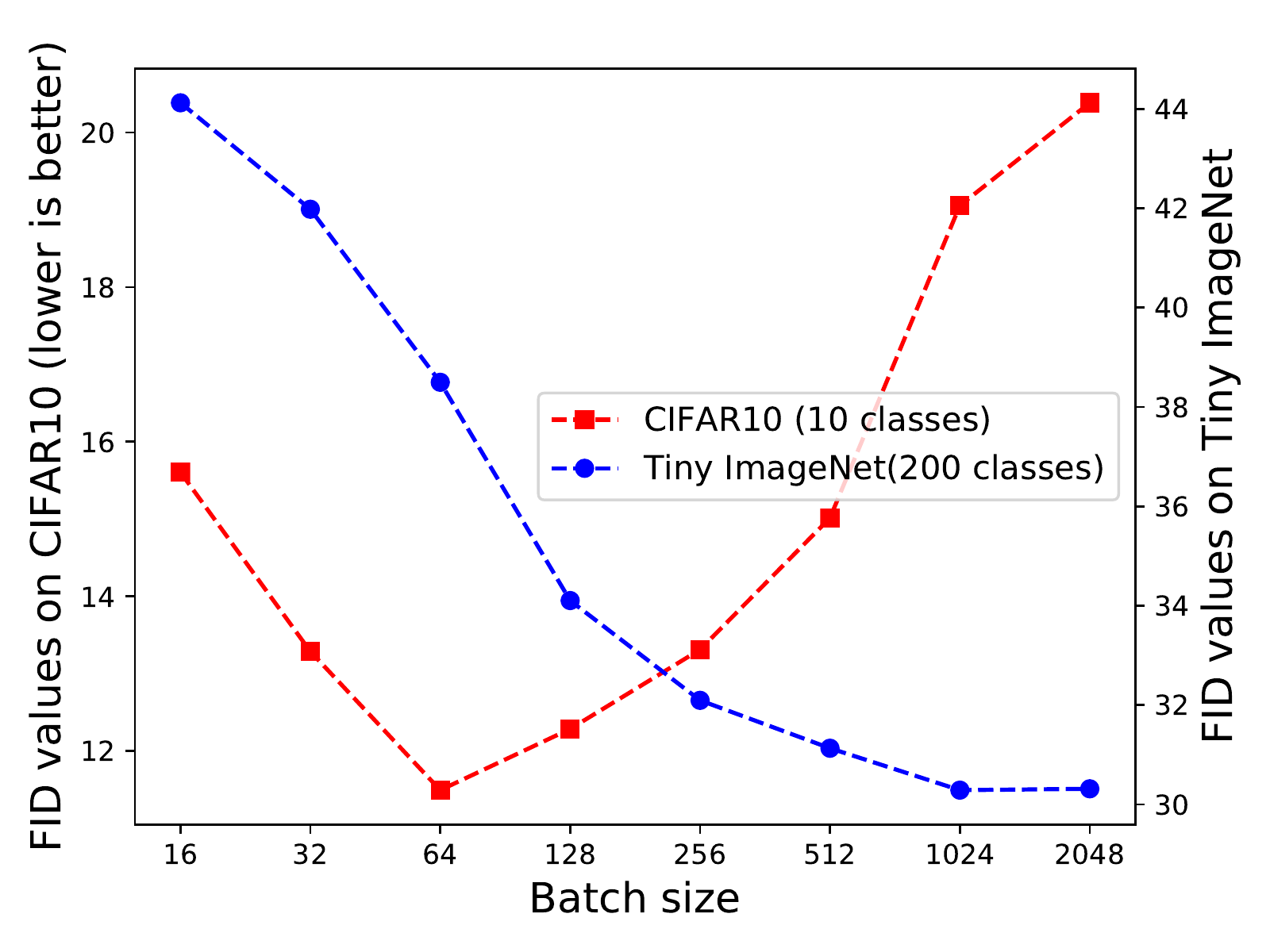}
         \caption{}
         \label{FigureA1b}
     \end{subfigure}
    \caption{(a) FID values of ContraGANs with different projection layers and embedding dimensionalities. (b) The change in FID values as the batch size increases. The experiments (a) and (b) are conducted using the setting (D).} \label{figure3}
\end{figure}
We study the effect of a projection layer $h: \mathbb{R}^k \longrightarrow \mathbb{S}^d$ that is introduced in Sec. 3.2. We change the types of the layer (linear vs. nonlinear) and increase the dimensionality of projected embeddings, $d$ on CIFAR10 dataset. Fig.~\ref{FigureA1a} shows the overview of FID values. All experiments are conducted using three different architectures: DCGAN, ResGAN, and BigGAN that are equipped with spectral normalization. We also run the experiments using three different random seeds and do not apply moving average update of the generator's weights. SNDCGAN with the liner projection layer projects latent features onto the 1,024 dimensional space. This configuration shows higher FID than the nonlinear counterpart, but ContraGANs with a linear projection layer generally give lower FIDs. Although GANs are known to need careful hyperparameter selection, our ContraGAN does not seem to be sensitive to the type and dimensionality of the projection layer.

Figure~\ref{FigureA1b} shows the change in FID values as the batch size increases. Experiments conducted by Brock~\etal~\cite{Brock2019LargeSG} have demonstrated that increasing the batch size enhances image generation performance on ImageNet dataset~\cite{Deng2009ImageNetAL}. However, as shown in Fig.~\ref{FigureA1b}, optimal batch sizes for CIFAR10 and Tiny ImageNet are 64 and 1,024, respectively. Based on these results, we can deduce that increasing batch size does not always give the best synthesis results. We presume that this phenomenon is related to the number of classes used for the training. 
\section{FID Implementations}
\begin{table}[h]
  \renewcommand\thetable{A\arabic{table}}
  \caption{Comparison of TensorFlow and PyTorch FID implementations.}
  \label{tableA10}
  \vspace{3.0mm}
  \centering
  \begin{tabular}{cccccccc}
    \toprule
     & \multicolumn{2}{c}{ContraGAN}\\
    FID implementation & CIFAR10~\cite{Krizhevsky2009LearningML} &  Tiny ImageNet~\cite{Tiny}\\
    \midrule
    TensorFlow~\cite{ttur2010inline} & 10.308 &  26.924  \\
    PyTorch~\cite{pytorchttur} & 10.304 &  27.131  \\
    \bottomrule
  \end{tabular}
\end{table}
FID is a widely used metric to evaluate the performance of a GAN model. Since calculating FID requires a pre-trained inception-V3 network~\cite{Szegedy2016RethinkingTI}, many implementations use Tensorflow~\cite{tensorflow2015-whitepaper} or PyTorch~\cite{NEURIPS2019_9015} libraries. Among them, the TensorFlow implementation~\cite{ttur2010inline} for FID measurement is widely used. We use the PyTorch implementation for FID measurement~\cite{pytorchttur}, instead. In this section, we show that the PyTorch-based FID implementation~\cite{pytorchttur} used in our work provides almost the same results as the TensorFlow implementation. The results are summarized in Table~\ref{tableA10}.
\section{Multiple Runs of the Stability Experiment}
\begin{figure}[t]
    \renewcommand\thefigure{A\arabic{figure}}
    \centering
    \includegraphics[width=0.31\linewidth]{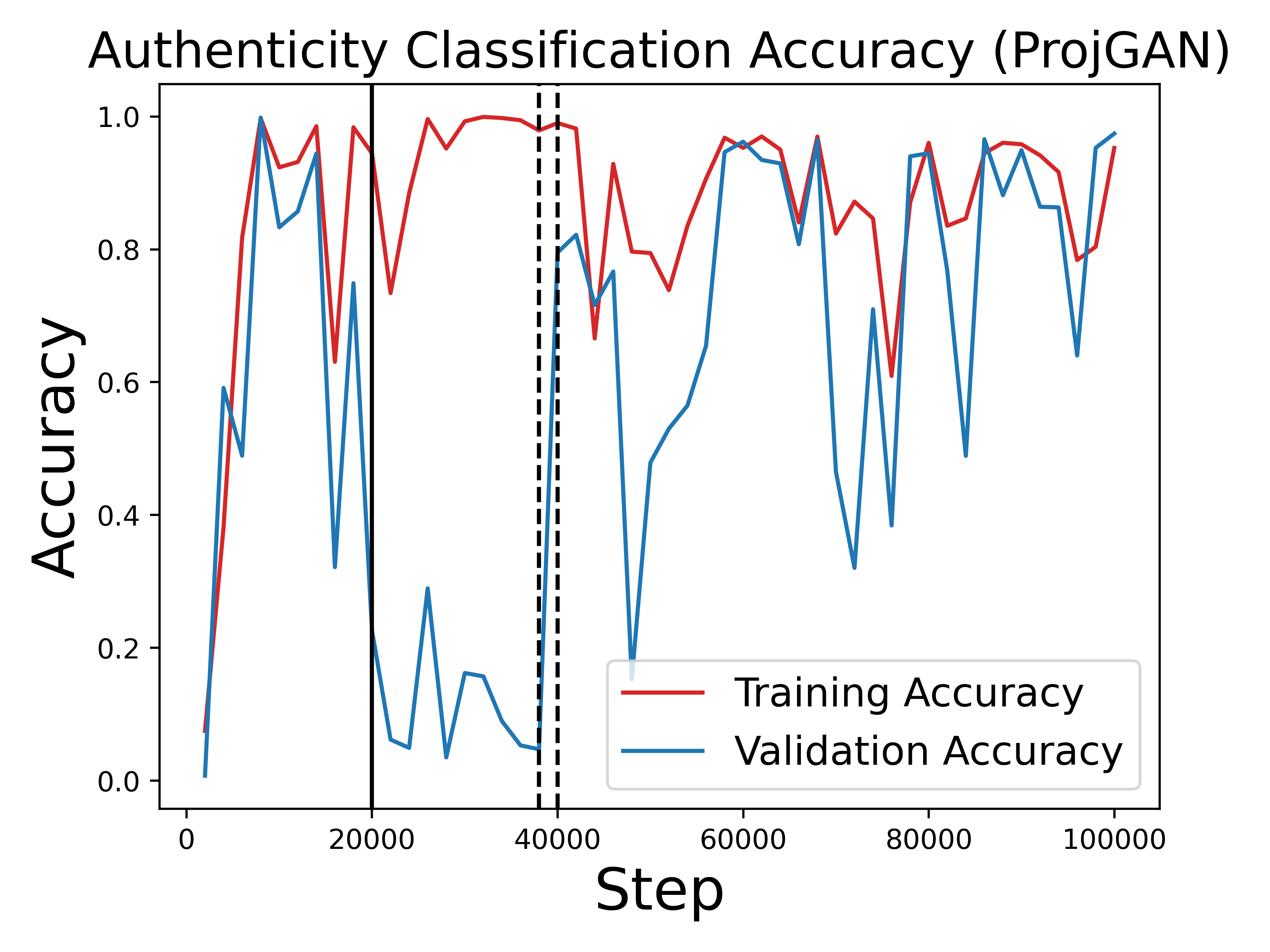}
    \includegraphics[width=0.31\linewidth]{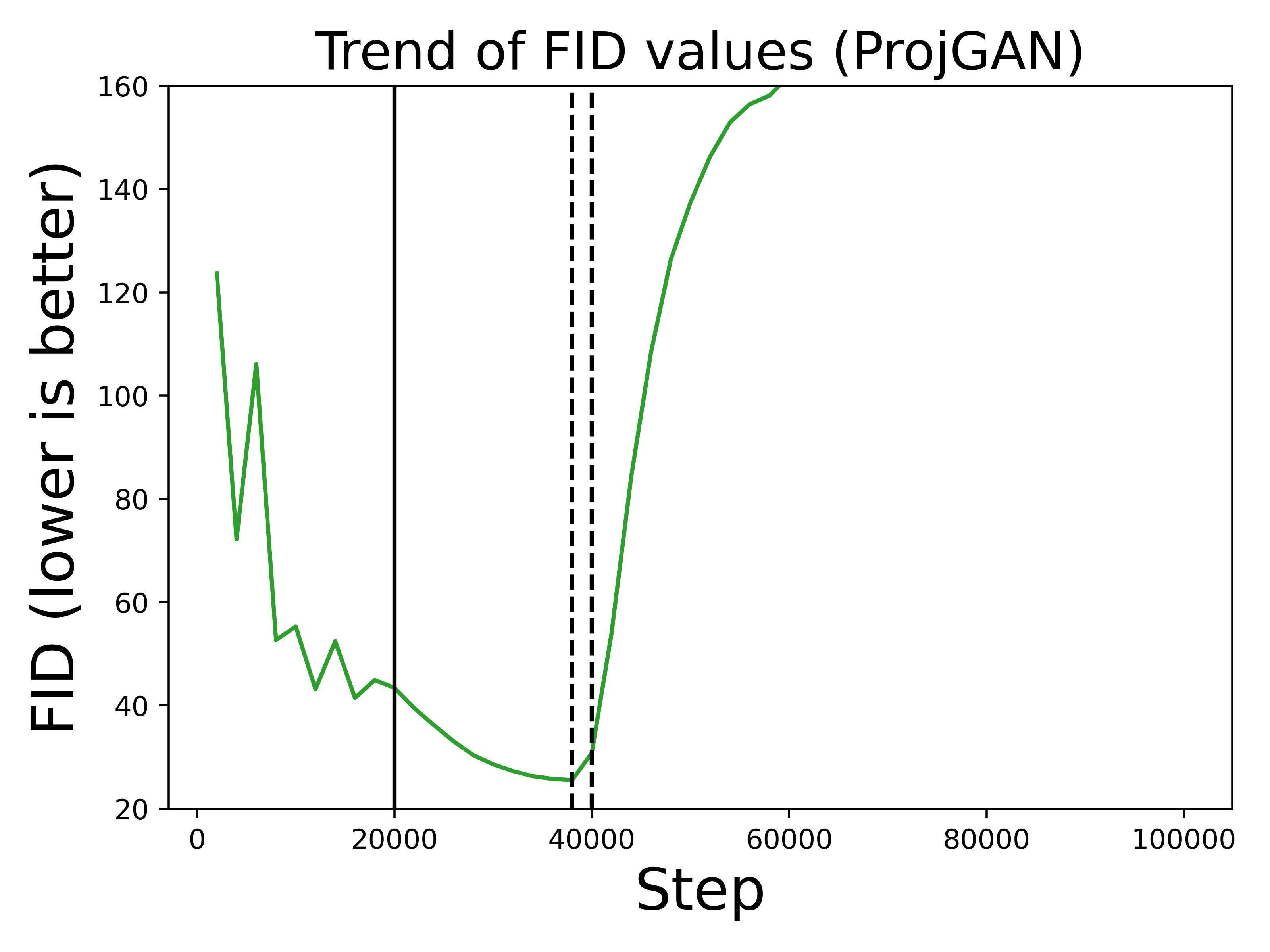}
    \includegraphics[width=0.31\linewidth]{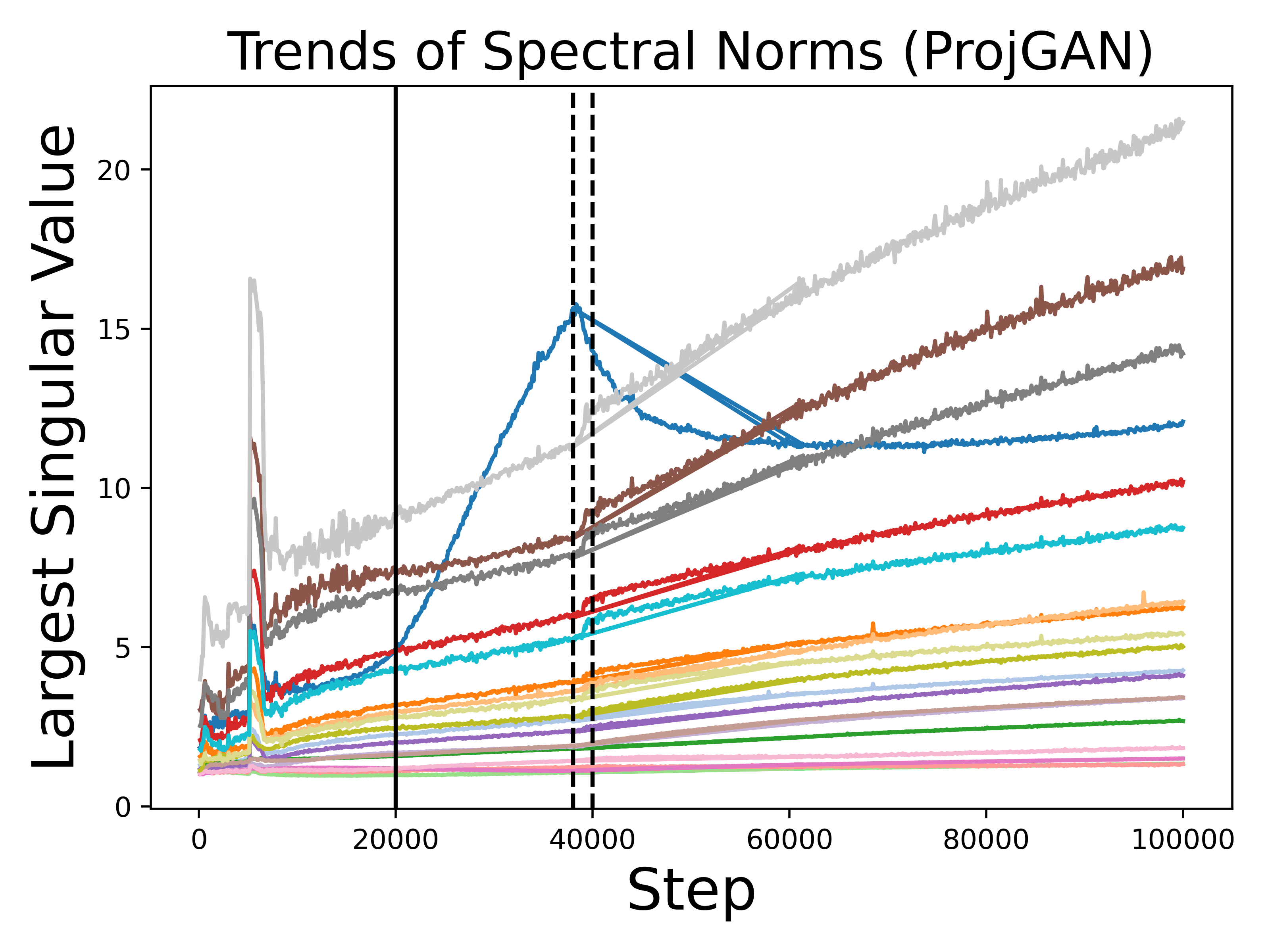}
    \includegraphics[width=0.31\linewidth]{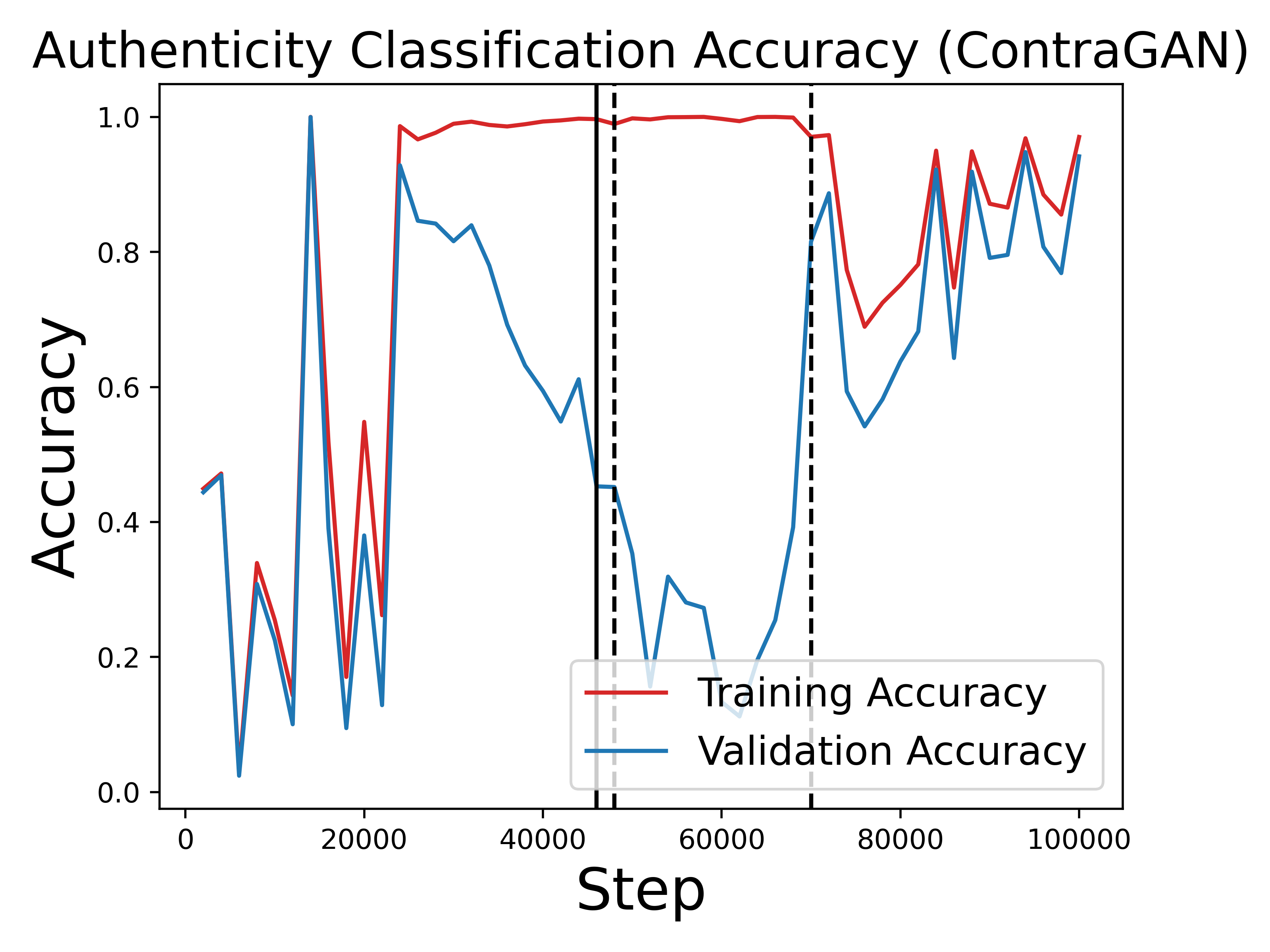}
    \includegraphics[width=0.31\linewidth]{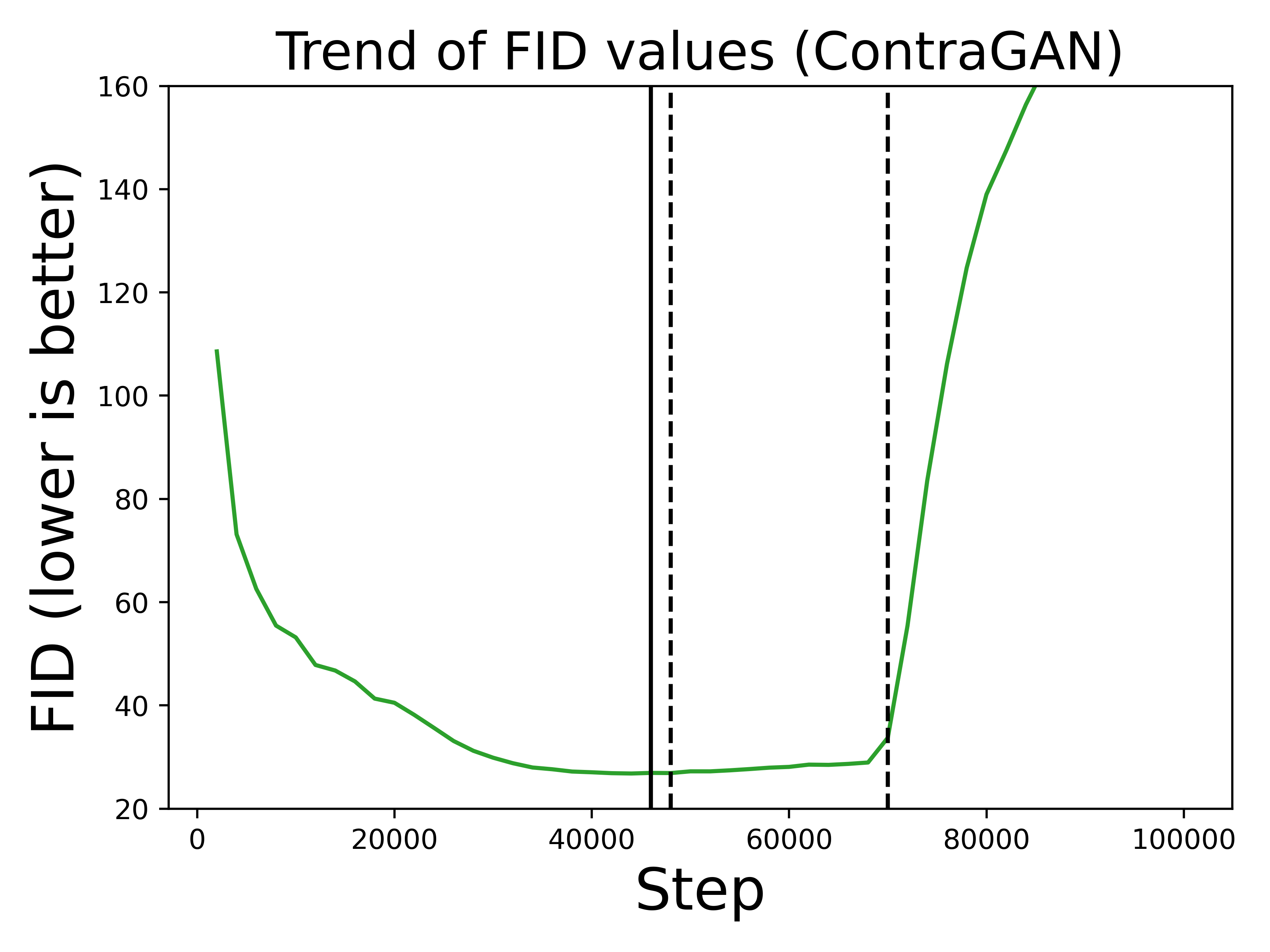}
    \includegraphics[width=0.31\linewidth]{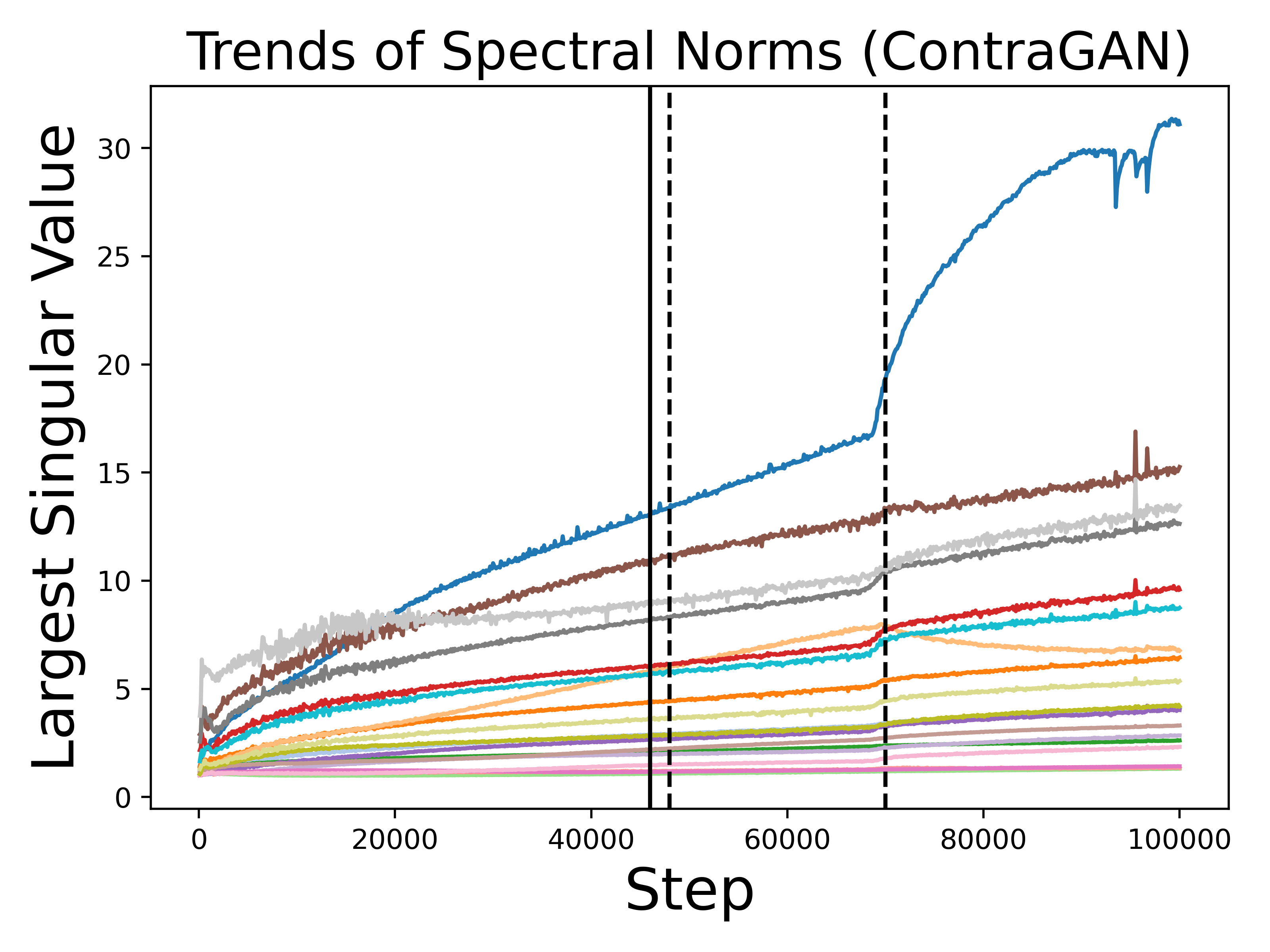}
    \includegraphics[width=0.31\linewidth]{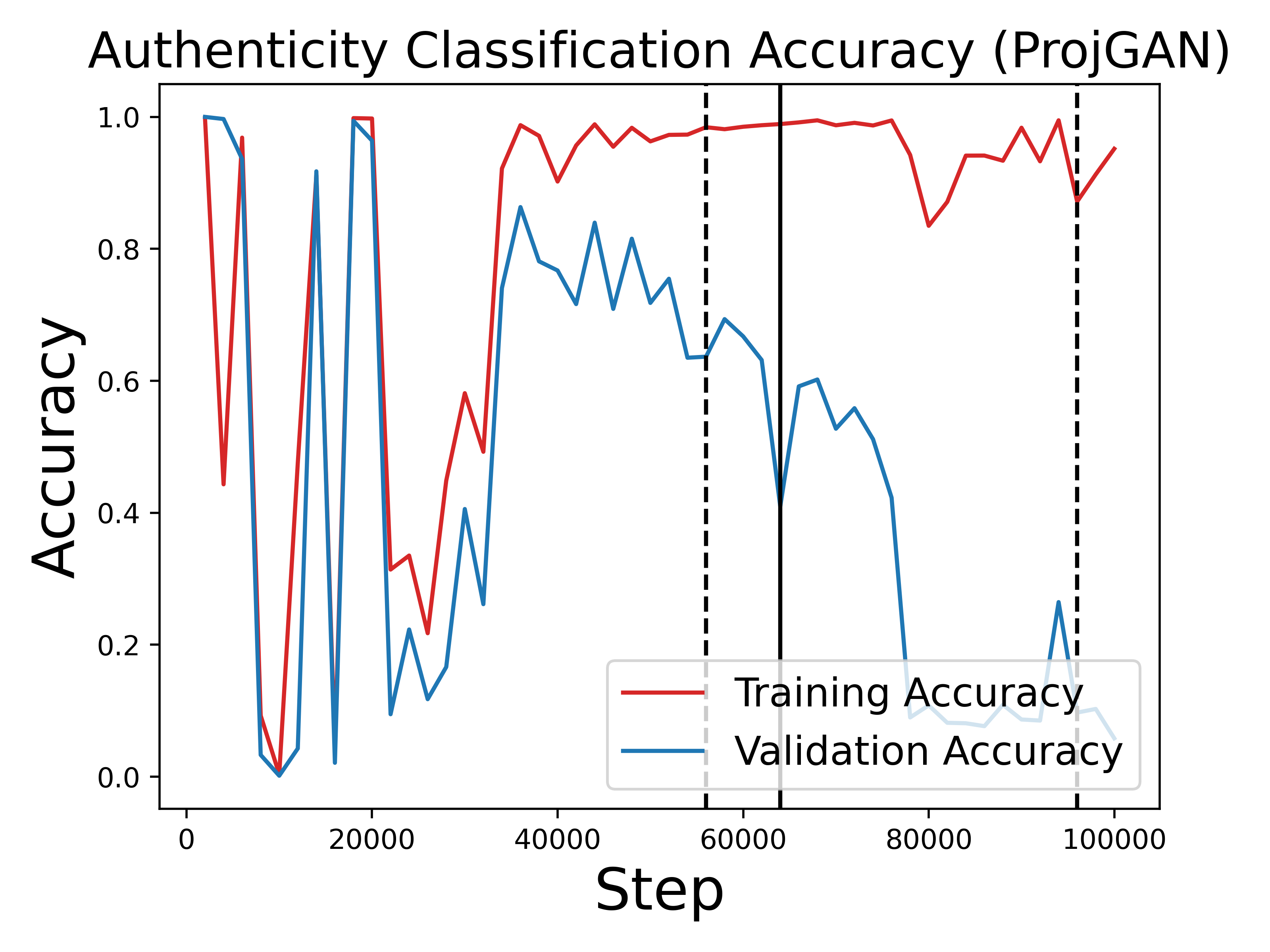}
    \includegraphics[width=0.31\linewidth]{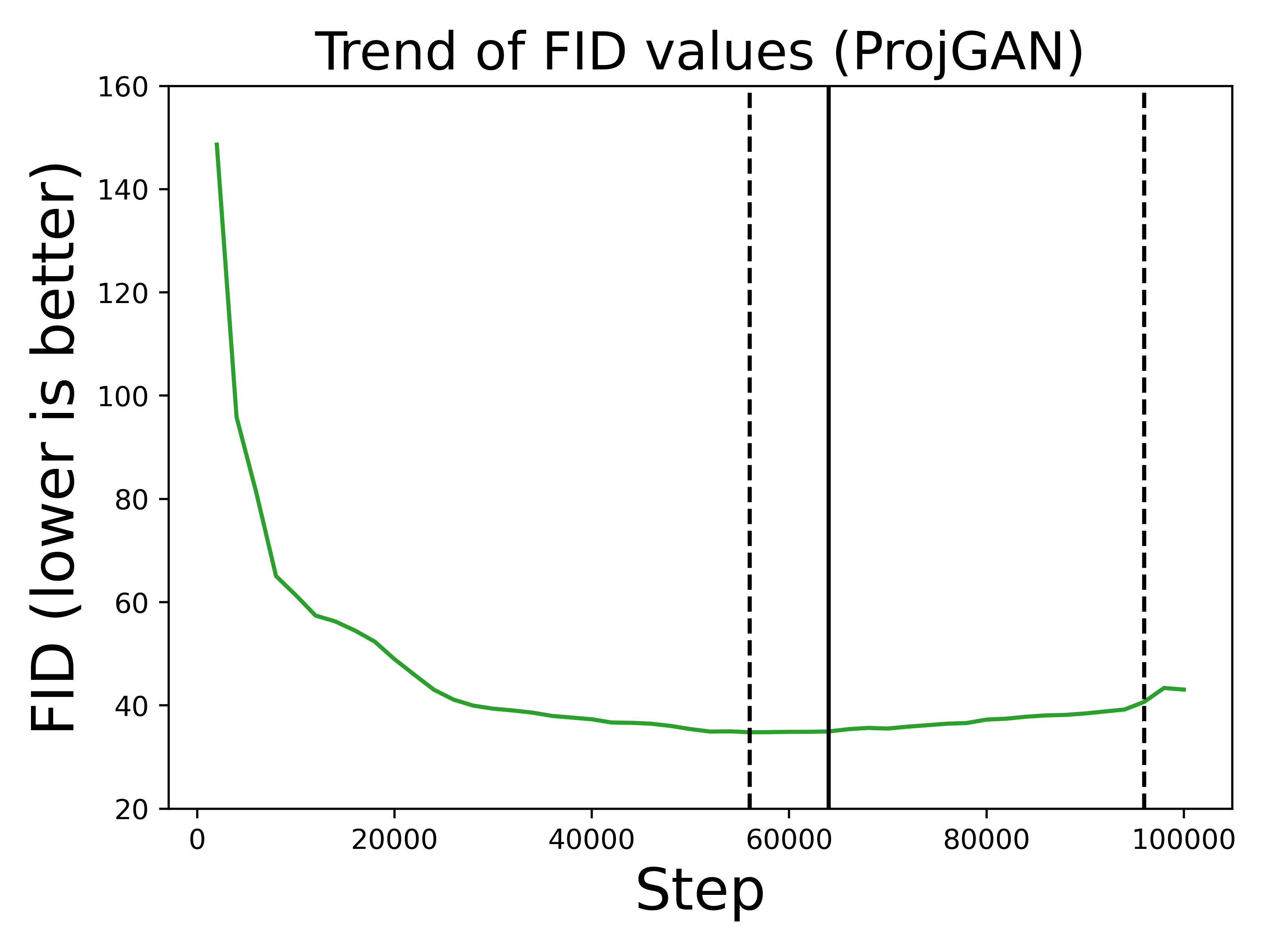}
    \includegraphics[width=0.31\linewidth]{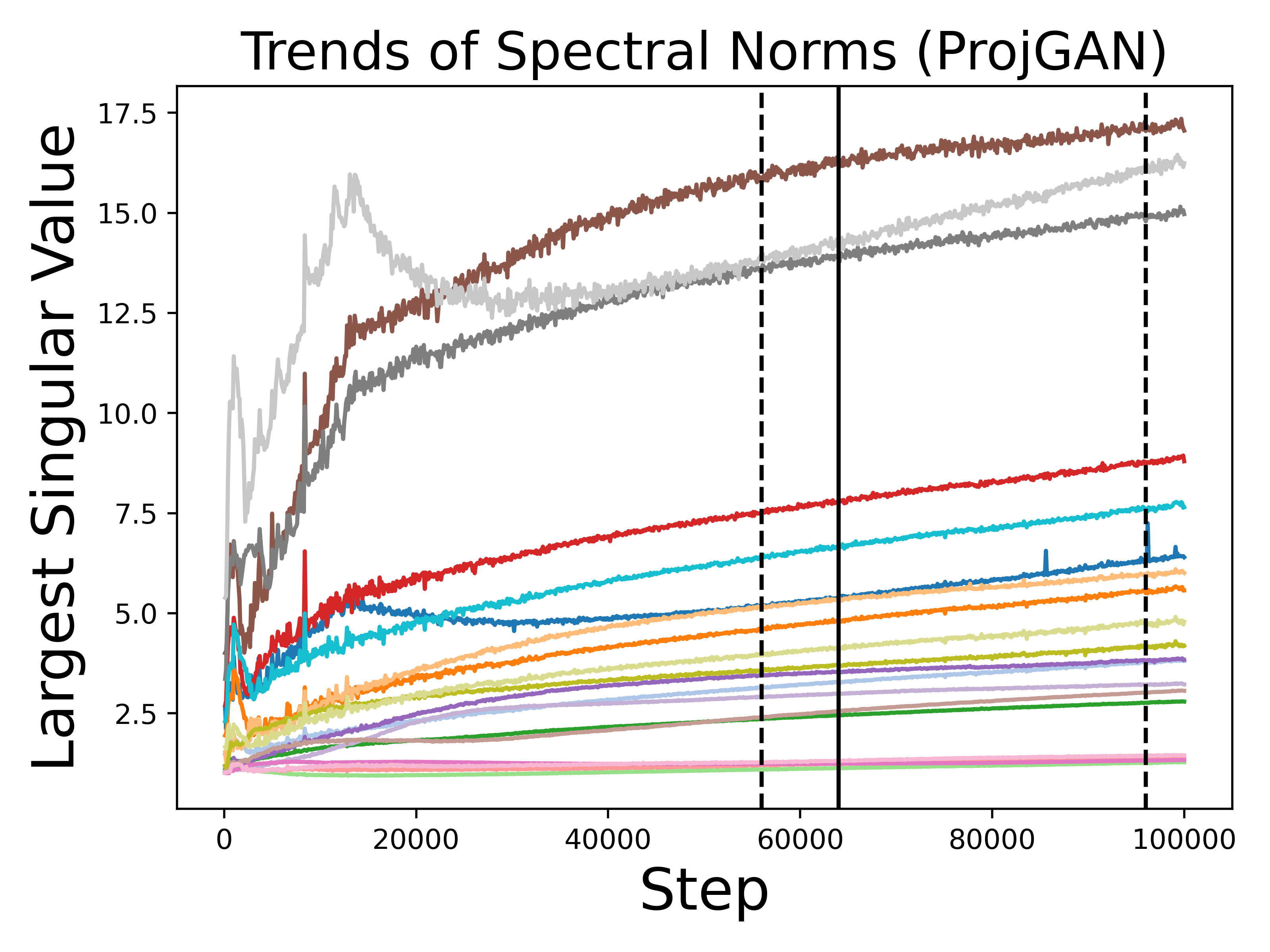}
    \includegraphics[width=0.31\linewidth]{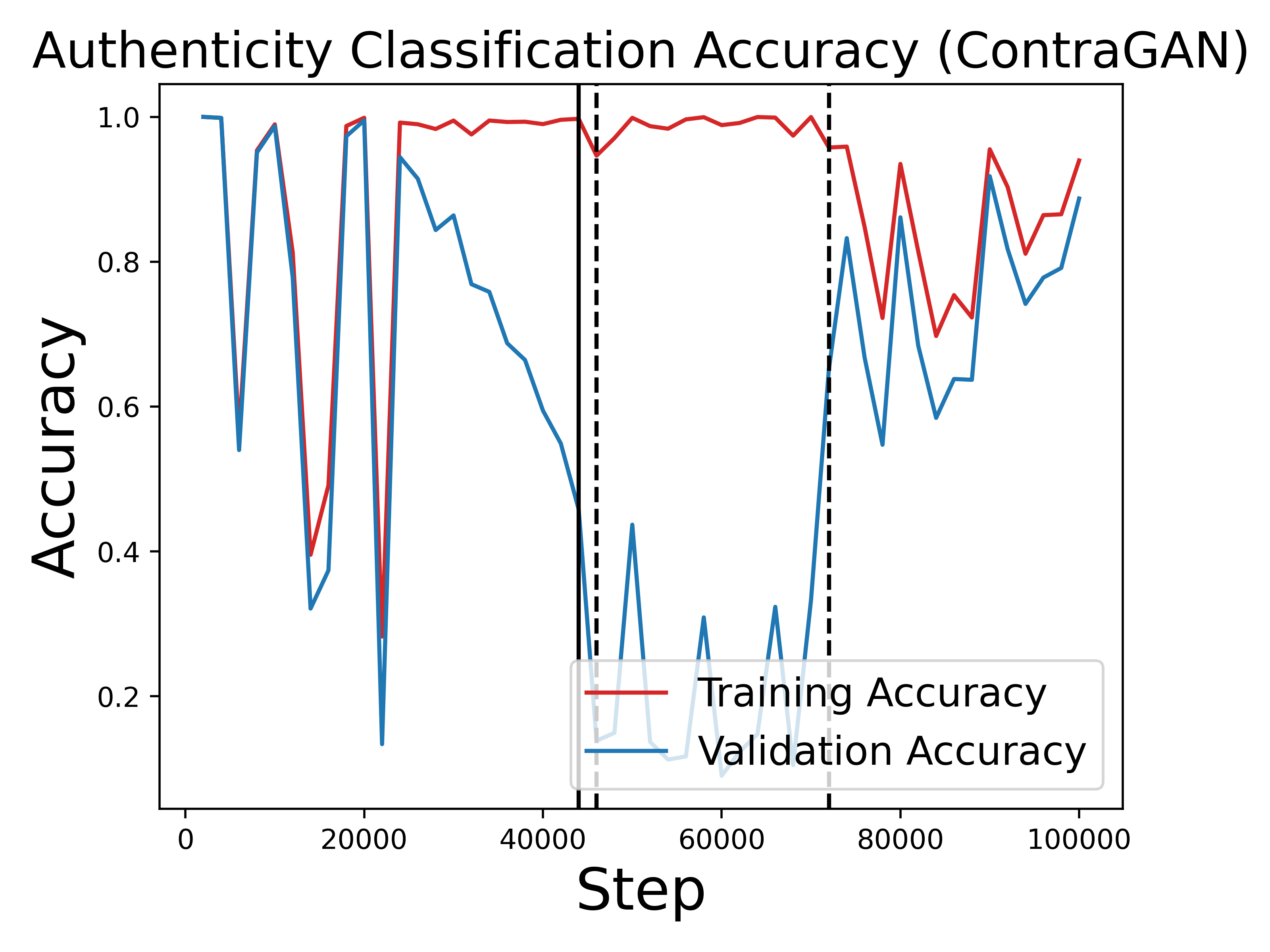}
    \includegraphics[width=0.31\linewidth]{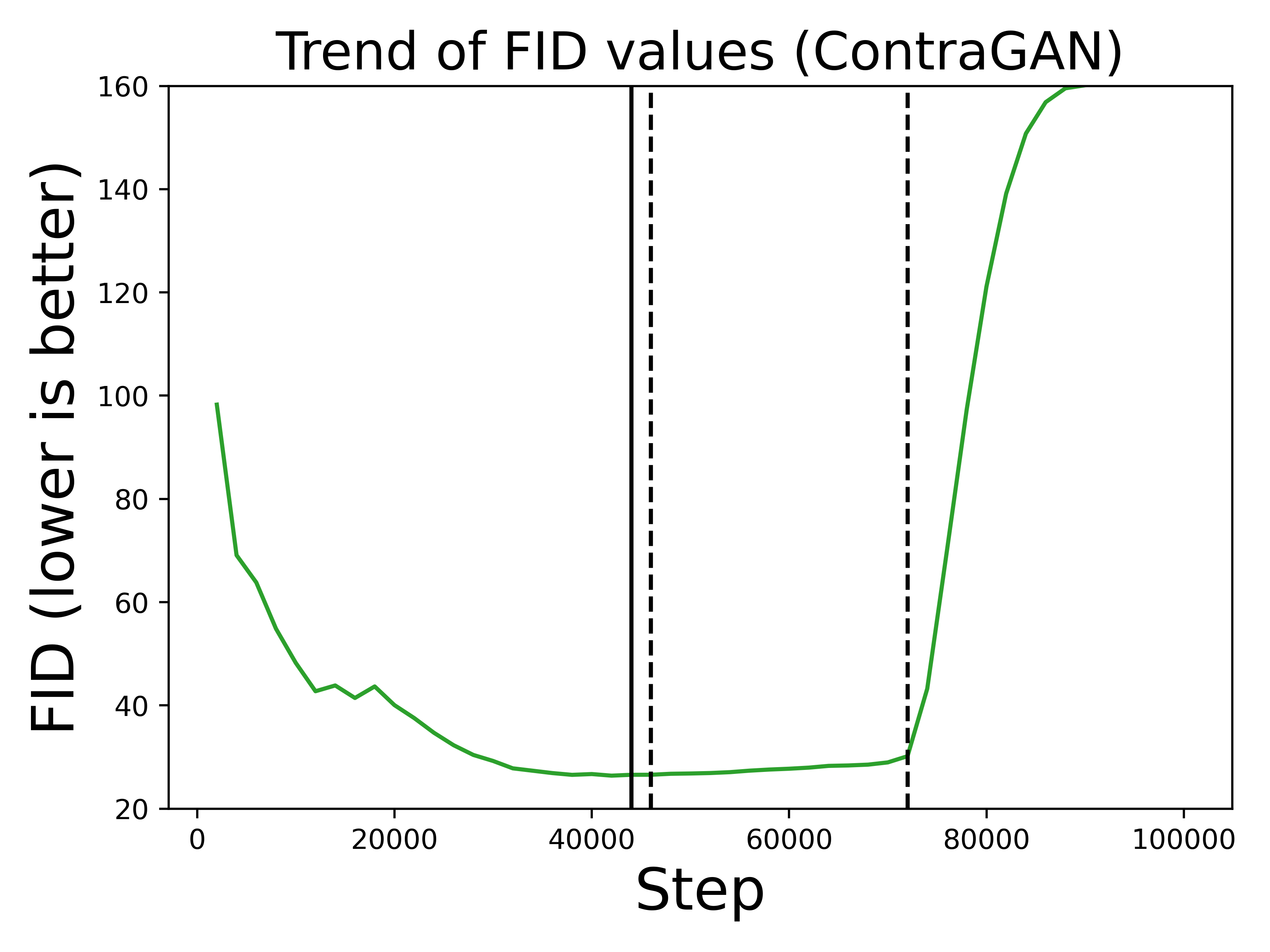}
    \includegraphics[width=0.31\linewidth]{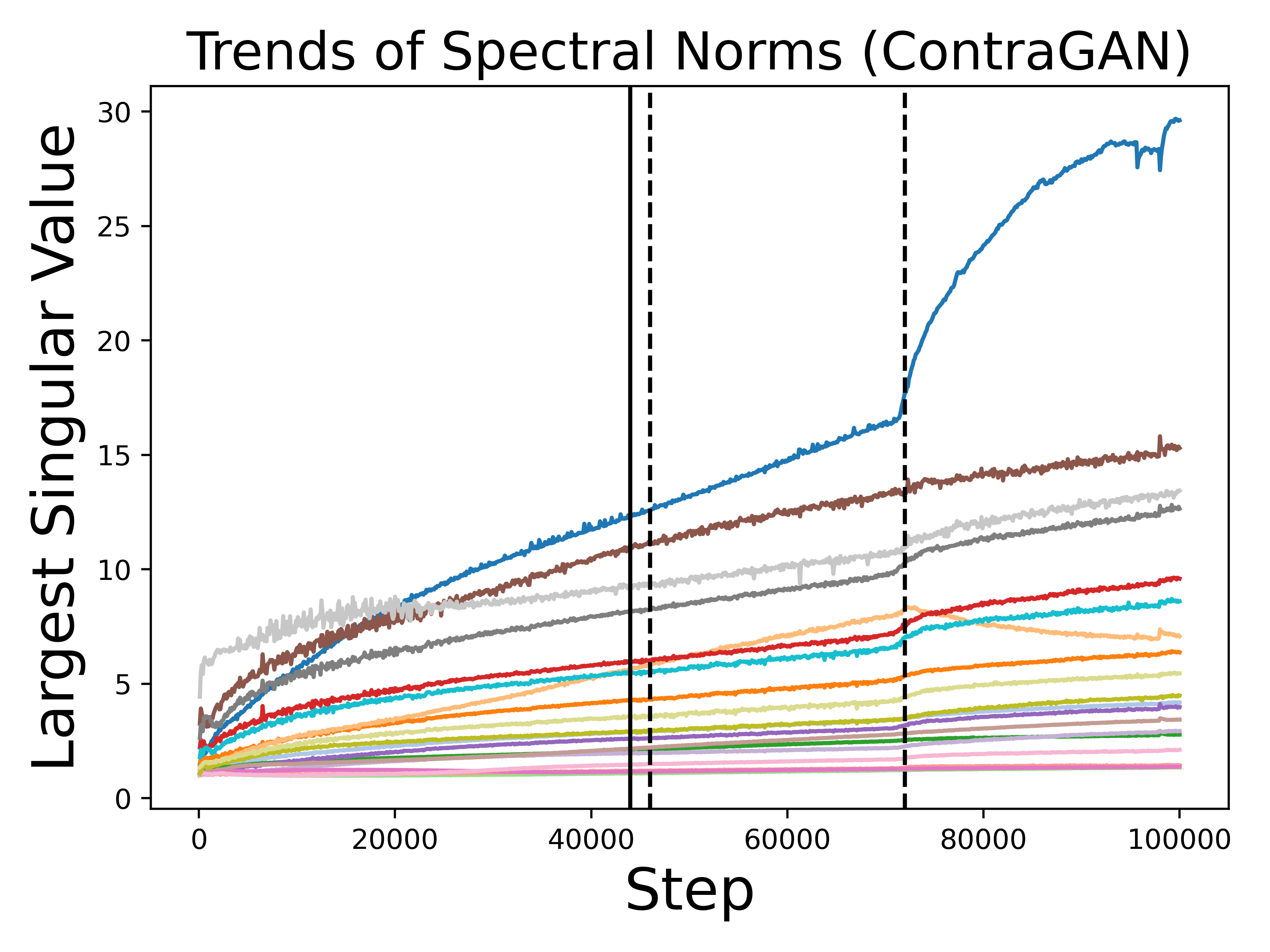}
    \caption{Authenticity classification accuracies on the training and validation datasets (left), trends of FID values (middle), and trends of the largest singular values of the discriminator's convolutional parameters (right). To specify the starting point where the difference between the training and validation accuracies is greater than 0.5, we use a solid black line. The first and second black dotted lines indicate when the performance is best and when training collapse occurs, respectively.}
    \label{fig:figa3}
\end{figure}

In this section, we provide the additional results of the stability test performed in Sec.~4.5 of the main paper. The third and fourth row of Fig.~\ref{fig:figa3} shows the another run from ProjGAN and ContraGAN.
\clearpage
As shown in the third row of Fig.~\ref{fig:figa3}, training collapse does not occur in training ProjGAN~\cite{Miyato2018cGANsWP}. However, the best FID value of the ProjGAN is 34.831, which is much higher than that of ContraGAN~($25\leq\text{FID}\leq27$). The above results show that ContraGAN is more robust to the overfitting and training collapse.
\section{Qualitative Results}
\begin{figure}[ht]
    \renewcommand\thefigure{A\arabic{figure}}
    \centering
    \includegraphics[scale=.48]{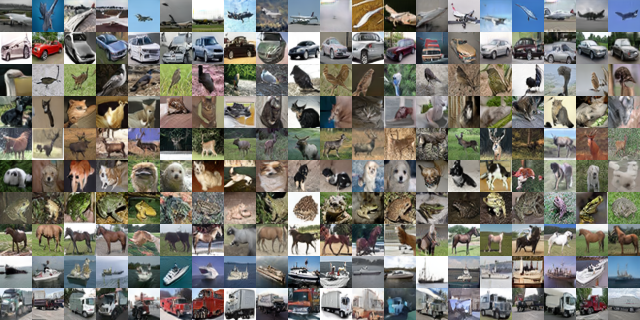}
    \caption{Examples generated by ACGAN~\cite{Odena2017ConditionalIS} trained on CIFAR10 dataset~\cite{Krizhevsky2009LearningML} (FID=11.111).} 
    \label{FigureA4a}
\end{figure}

\begin{figure}[ht]
    \renewcommand\thefigure{A\arabic{figure}}
    \centering
    \includegraphics[scale=.48]{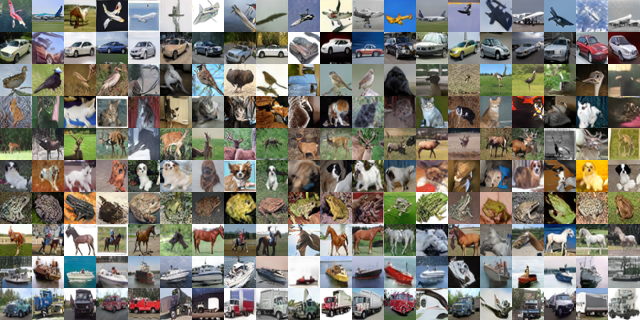}
    \caption{Examples generated by ProjGAN~\cite{Miyato2018cGANsWP} on CIFAR10 dataset~\cite{Krizhevsky2009LearningML} (FID=10.933).} 
    \label{FigureA4b}
\end{figure}

\begin{figure}[ht]
    \renewcommand\thefigure{A\arabic{figure}}
    \centering
    \includegraphics[scale=.48]{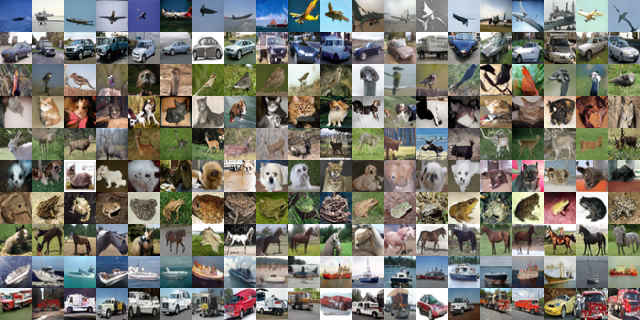}
    \caption{Examples generated by ContraGAN (Ours) on CIFAR10 dataset~\cite{Krizhevsky2009LearningML} (FID=10.188).} 
    \label{FigureA4c}
\end{figure}

\begin{figure}[ht]
    \renewcommand\thefigure{A\arabic{figure}}
    \centering
    \includegraphics[scale=.6]{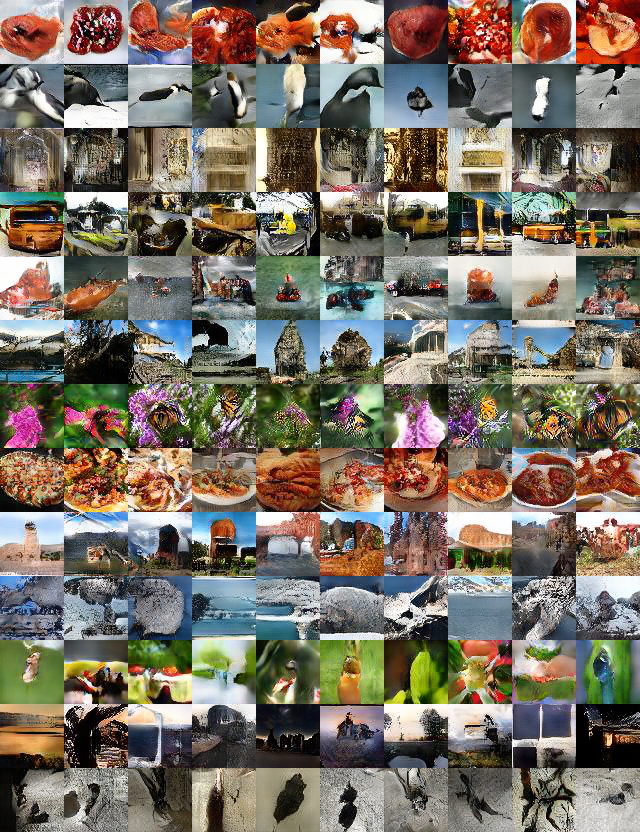}
    \caption{Examples generated by ProjGAN~\cite{Miyato2018cGANsWP} on Tiny ImageNet dataset~\cite{Tiny} (FID=34.090).} 
    \label{FigureA4d}
\end{figure}

\begin{figure}[ht]
    \renewcommand\thefigure{A\arabic{figure}}
    \centering
    \includegraphics[scale=.6]{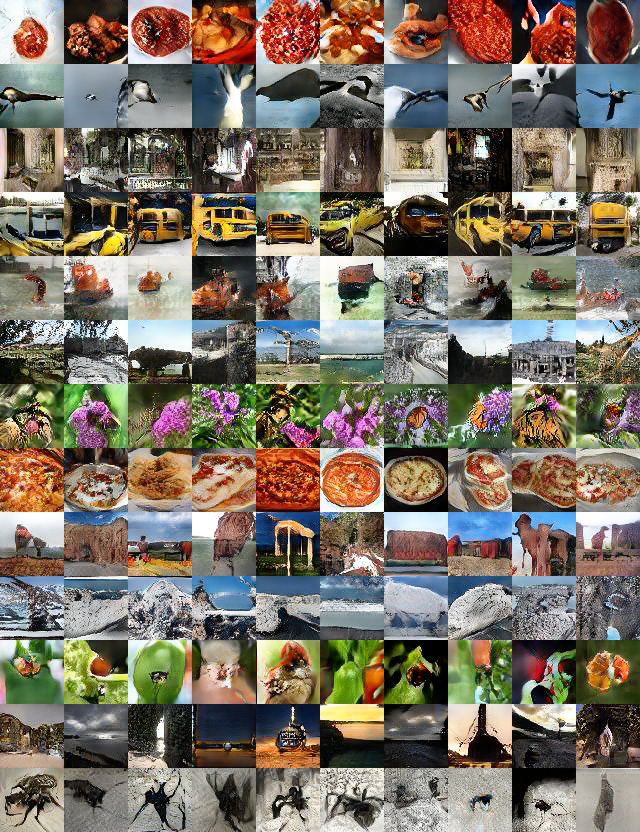}
    \caption{Examples generated by ContraGAN (Ours) on Tiny ImageNet dataset~\cite{Tiny}~(FID=30.286).} 
    \label{FigureA4e}
\end{figure}

\begin{figure}[ht]
    \renewcommand\thefigure{A\arabic{figure}}
    \centering
    \includegraphics[scale=.5]{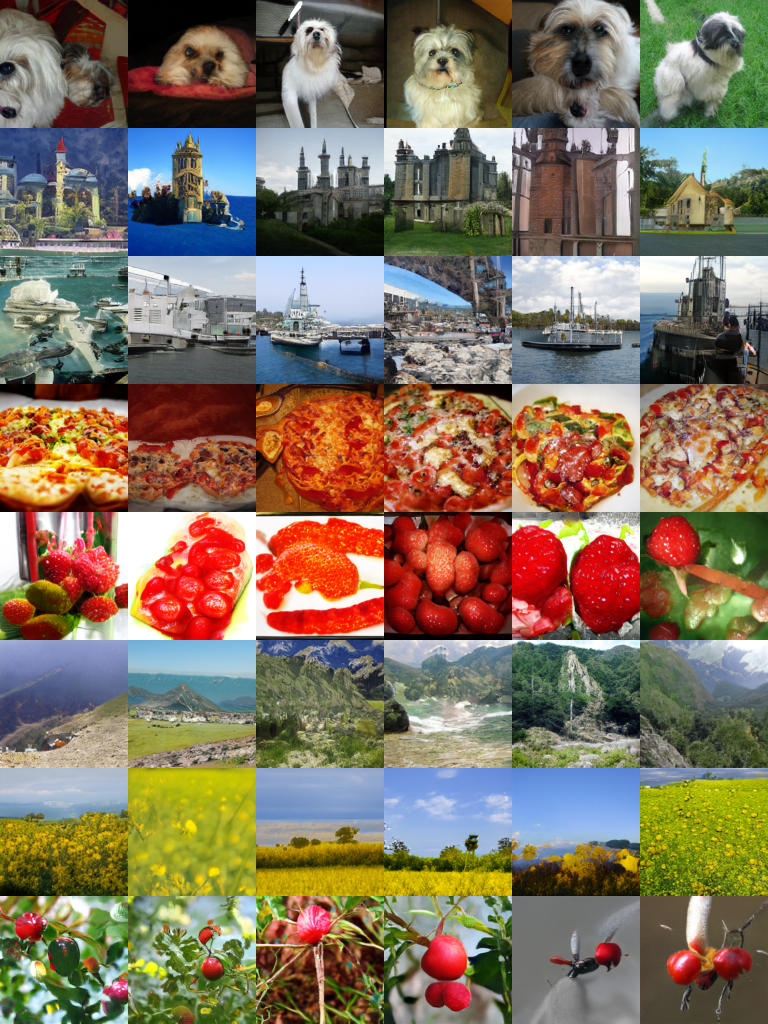}
    \caption{Examples generated by ProjGAN~\cite{Miyato2018cGANsWP} on ImageNet dataset~\cite{Deng2009ImageNetAL} (FID=21.072).} 
    \label{FigureA4f}
\end{figure}

\begin{figure}[ht]
    \renewcommand\thefigure{A\arabic{figure}}
    \centering
    \includegraphics[scale=.5]{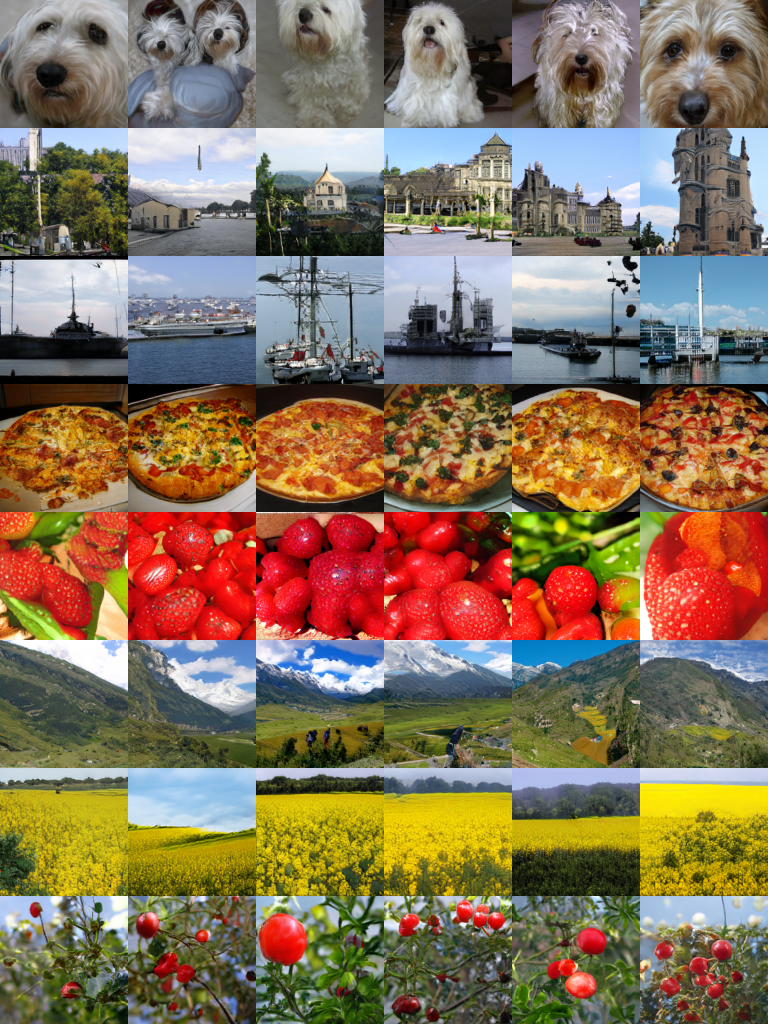}
    \caption{Examples generated by ContraGAN (Ours) on ImageNet dataset~\cite{Deng2009ImageNetAL}~(FID=19.443).} 
    \label{FigureA4g}
\end{figure}

This section presents images generated by various conditional image generation frameworks. Fig.~\ref{FigureA4a}, \ref{FigureA4b}, and \ref{FigureA4c} show the synthesized images using CIFAR10 dataset. Fig.~\ref{FigureA4d} and \ref{FigureA4e} show the synthesized images using Tiny ImageNet dataset. Fig.~\ref{FigureA4f} and \ref{FigureA4g} show the generated images using ImageNet dataset. As shown in Fig.~\ref{FigureA4e} and \ref{FigureA4g}, our approach can achieve favorable FID compared to the other baseline approaches.

\end{document}